\documentclass{applemlr}

\usepackage{amsmath}
\usepackage{enumerate}
\usepackage{algorithm}
\usepackage{algpseudocode}
\usepackage{amsfonts}
\usepackage{amsthm}
\usepackage{cleveref}
\usepackage{diagbox}
\usepackage{colortbl}
\usepackage{amssymb}
\usepackage{xspace}
\usepackage{wrapfig}
\usepackage{adjustbox}
\usepackage{tabularx}
\usepackage{booktabs}
\usepackage{mathtools}
\usepackage{tikz}
\usepackage{enumitem}
\usepackage{silence}
\usepackage{dsfont}
\usepackage[table]{xcolor}
\usepackage[dvipsnames]{xcolor}
\usepackage{multirow}
\usepackage{makecell}
\usepackage{xfakebold}

\usepackage{amsmath,amsfonts,bm}

\def\eqref#1{equation~\ref{#1}}

\def\1{\bm{1}}

\DeclareMathAlphabet{\mathsfit}{\encodingdefault}{\sfdefault}{m}{sl}
\SetMathAlphabet{\mathsfit}{bold}{\encodingdefault}{\sfdefault}{bx}{n}

\setcitestyle{numbers,square,citesep={,}}

\crefname{appsec}{appendix}{appendices}
\Crefname{appsec}{Appendix}{Appendices}

\definecolor{textgray}{HTML}{6E6E73}
\usetikzlibrary{positioning, calc}
\usetikzlibrary{decorations.pathmorphing}

\makeatletter
\patchcmd{\wrong@fontshape}{\@gobbletwo}{}{}{}
\makeatother
\numberwithin{equation}{section}
\makeatletter
\AtBeginDocument{
  \urlstyle{sf}
  
}
\makeatother

\definecolor{light}{RGB}{125, 125, 125}
\crefname{tcb@cnt@pbox}{code}{code}
\Crefname{tcb@cnt@pbox}{Code}{Code}
\crefname{assumption}{assumption}{assumption}
\Crefname{assumption}{Assumption}{Assumptions}

\newtcolorbox[auto counter]{pbox}[2][]{
  colback=white,
  title=Code~\thetcbcounter: #2,
  #1,fonttitle=\sffamily,
  fontupper=\sffamily,
  arc=2pt,
  colframe=bgcolor,
  coltitle=fgcolor,
  colbacktitle=bgcolor,
  toptitle=0.25cm,
  bottomtitle=0.125cm
}

\makeatletter
\newcommand\applefootnote[1]{%
  \begingroup
  \renewcommand\thefootnote{}%
  \renewcommand\@makefntext[1]{\noindent##1}%
  \footnote{#1}%
  \addtocounter{footnote}{-1}%
  \endgroup
}
\makeatother

\definecolor{cverbbg}{gray}{0.90}

\usepackage[utf8]{inputenc} 
\usepackage[T1]{fontenc}    
\usepackage{hyperref}       
\usepackage{url}            
\usepackage{booktabs}       
\usepackage{amsfonts}       
\usepackage{nicefrac}       
\usepackage{microtype}      
\usepackage{xcolor}         

\usepackage{graphicx}
\usepackage{amsmath}
\usepackage{amssymb}
\usepackage{float}
\usepackage{tikz}
\usepackage{xspace}
\usepackage{cleveref}
\usepackage{colortbl}
\usepackage{arydshln}
\usepackage{titletoc} 
\usepackage{tcolorbox}
\usepackage{wrapfig}
\usepackage{multirow}
\newcommand{\method}{\textsc{Weblica}\xspace}

\newcommand{\methodsynth}{\textsc{Weblica-Synth}\xspace}
\newcommand{\methodcache}{\textsc{Weblica-Cache}\xspace}

\newcommand{\modelnameeightb}{\textsc{Weblica-8B}\xspace}
\newcommand{\modelnamefourb}{\textsc{Weblica-4B}\xspace}
\newcommand{\modelnametwob}{\textsc{Weblica-2B}\xspace}

\makeatletter
\newcommand{\fakesc}[1]{\@fakesc#1\@nil}
\def\@fakesc#1#2\@nil{#1\scalebox{0.78}[0.78]{\MakeUppercase{#2}}}
\makeatother
\newcommand{\methodfakesc}{\fakesc{Weblica}\xspace}
  
\title{\methodfakesc: Scalable and Reproducible Training Environments for Visual Web Agents}

\author{Oğuzhan Fatih Kar}
\author{Roman Bachmann}
\author{Yuanzheng Gong}
\author{Anders Boesen Lindbo Larsen}
\author{Afshin Dehghan}

\affiliation{Apple}

\abstract{
The web is complex, open-ended, and constantly changing, making it challenging to scale training data for visual web agents. Existing data collection attempts remain limited to   
  offline trajectories for supervised fine-tuning or a handful of simulated environments for RL training, thus failing to capture web diversity. We propose \methodfakesc (Web Replica), a
  framework for constructing reproducible and scalable web environments. Our framework leverages 1) HTTP-level caching to capture and replay stable visual states while preserving   
  interactive behavior and 2) LLM-based environment synthesis grounded in real-world websites and core web navigation skills. Using this framework, we scale RL training to thousands
   of diverse environments and tasks. Our best model, \methodfakesc-8B, outperforms open-weight baselines of similar size across multiple web navigation benchmarks while using fewer inference steps, scales favorably with additional test-time compute, and is competitive with API models.}

\date{\sffamily\today}

\begin{document}

\maketitle

\section{Introduction}
\label{sec:intro}

Large language models are increasingly capable as autonomous agents in domains like coding~\citep{anthropic2025claudecode, openaicodex, wang2024openhands, hui2024qwen2}, mathematics~\citep{hubert2025olympiad,novikov2025alphaevolve}, and computer use~\citep{anthropic2024claudecomputeruse,openaicomputeruse,geminicomputeruse,qin2025ui}. This progress is driven by the availability of large-scale, high-quality training data. Web navigation has emerged as a recent focus toward building agents that autonomously navigate the web to solve tasks ranging from information retrieval to form filling to online shopping. These agents aim to complete multi-step workflows that currently require manual effort, representing a significant step toward personalized digital assistants.

\begin{figure}[h]
    \centering
    \includegraphics[width=0.64\linewidth]{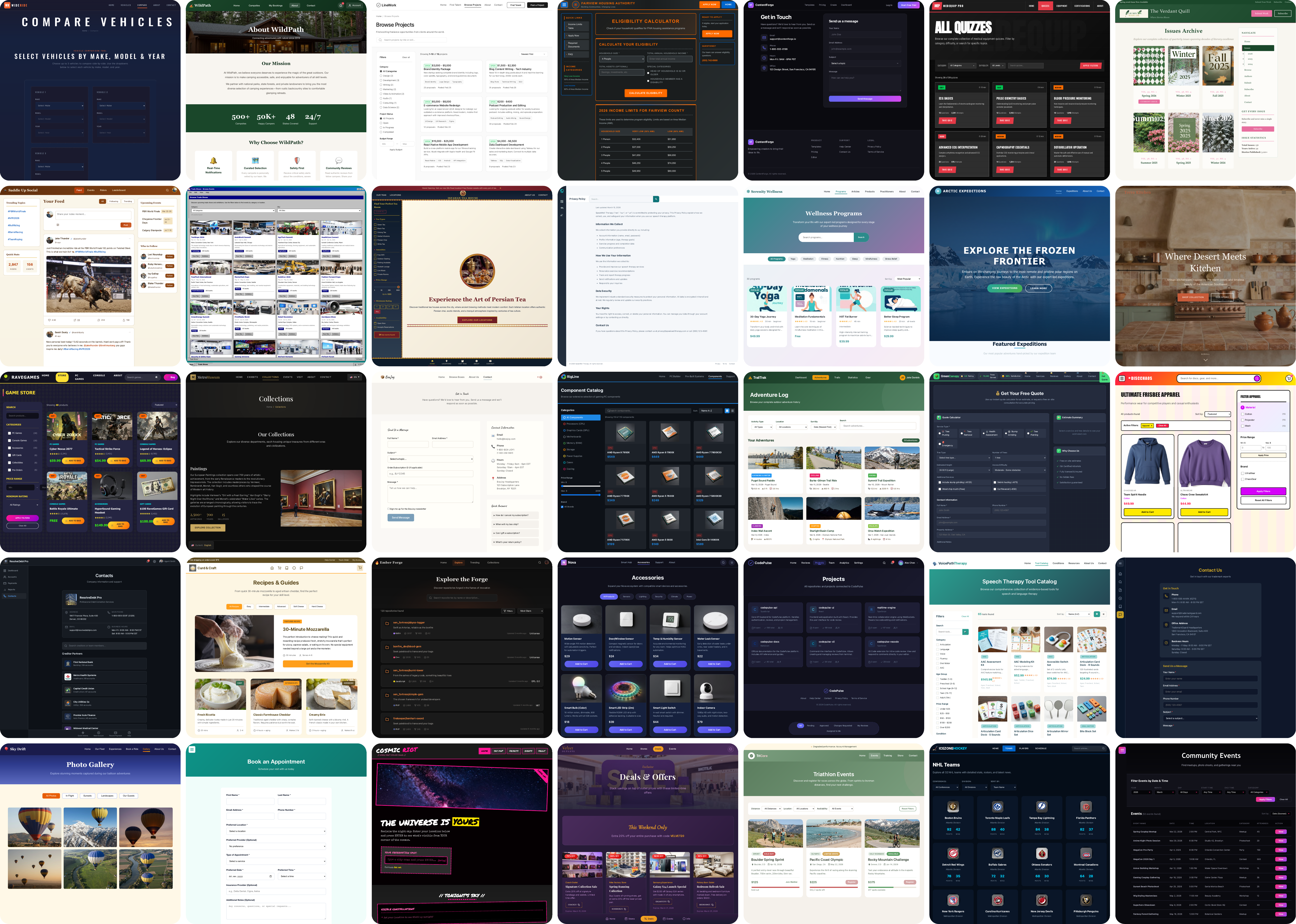}\hfill
    \includegraphics[width=0.34\linewidth]{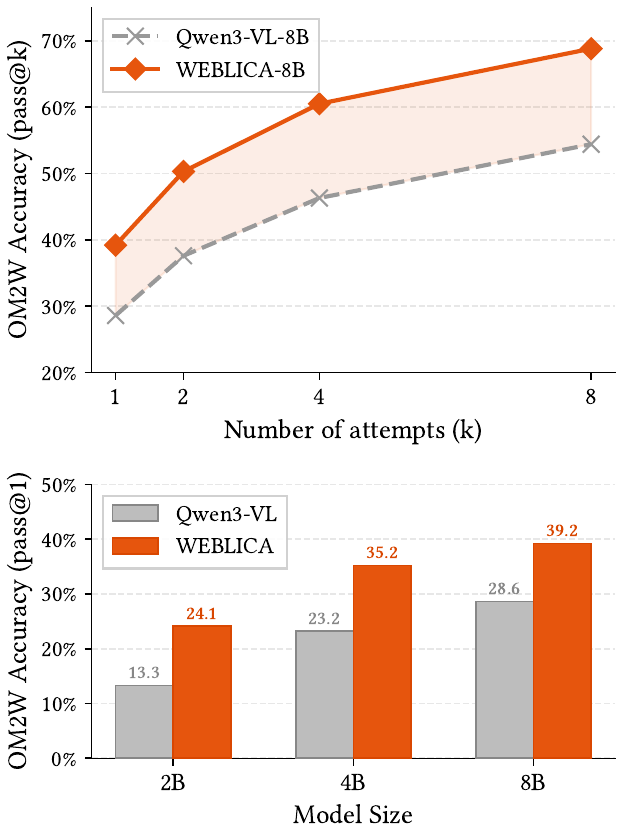}
    \caption{Scaling training data for visual web agents is challenging due to the complex, dynamic, and open-ended nature of the web. \textbf{Left:} We synthetically generate web environments in a fully automated and scalable manner, spanning a broad set of capabilities like navigation, form filling, filtering, date picking, and more. Alongside caching real websites, these create a fully offline alternative to the live web training. \textbf{Right:} Training on these environments improves performance across multiple web navigation benchmarks (Online-Mind2Web~\citep{xue2025illusion} shown here), scaling with both test-time compute~(top) and model size~(bottom).}       
    \label{fig:pullfigure}
  \end{figure} 

However, progress on building web agents has been slower, primarily due to the difficulty of scaling training data and environments to capture the complexity of the web. Recent attempts at data generation include collecting offline trajectories as demonstration data for supervised fine-tuning~\citep{wang2025opencua,awadallah2025fara,gupta2026molmoweb}, offering limited support to handle the stochastic nature of the web due to lack of interaction. As an alternative, building simulated web environments offers interaction, but they commonly cover only a handful of manually defined domains~\citep{zhou2023webarena, koh2024visualwebarena}, limiting generalization. While training directly on live websites could alleviate that, it suffers from brittleness due to timeouts and bot detection, making training unstable. In addition, the live web is constantly evolving and slow to interact with, making carefully controlled ablations and fully reproducible training difficult. This raises a natural question: \textit{how to scale interactive web environments while maintaining reproducibility?}

We propose \method, a framework for constructing reproducible and scalable web environments to train visual web agents. Our framework introduces two complementary mechanisms. First, we develop an HTTP-level caching system that records and replays real website interactions, capturing stable visual states while preserving interactive behavior. This enables reproducible training on diverse real-world websites without the brittleness of live web training, though it is limited to domains where stable recordings can be obtained. Second, we present an LLM-based environment synthesis pipeline that generates interactive web environments grounded in real websites and core web navigation skills (e.g., form submission, authentication flows, dynamic search), enabling scaling to broader domains at the cost of a potential sim-to-real gap. All environments are served locally, eliminating network latency and enabling fast training. Together, these approaches provide diverse, reproducible environments at scale.

Using this framework, we scale training to thousands of diverse environments and web navigation tasks. We fine-tune models
  from the Qwen3-VL~\citep{yang2025qwen3} family, which operate purely on screenshots without requiring set-of-marks annotations or DOM access, as these can hurt generalization due to the web's inconsistent underlying structure~\citep{yutori2025bitter}. We study the effect of training stages and environment composition, and analyze how performance
  scales with model size and test-time compute. Our 8B model achieves strong results across multiple web navigation benchmarks. On Online-Mind2Web~\citep{xue2025illusion}, it reaches 39.2\% pass@1 with only 30 steps, outperforming open-weight models that use 3$\times$ more steps, and improves further with additional test-time compute.

\section{Related Work}
\label{sec:related}

\paragraph{Building Web Agents.}                                                                                                                
  Early web agents relied on text-only language models that process structured representations such as accessibility trees or DOM              
  elements~\citep{yao2022webshop,deng2023mind2web,zhou2023webarena,gur2023real}. Later work adopted vision-language models (VLMs) to ground actions
  visually~\citep{koh2024visualwebarena,he2024webvoyager,hong2024cogagent}. Since early VLMs had limited grounding capabilities, initial        
  approaches augmented screenshots with set-of-marks~\citep{yang2023set} overlays. These overlay numbered bounding boxes on interactive elements
   to simplify action prediction. However, this introduces dependencies on accurate element detection and adds visual clutter that does not
  reflect natural web perception~\citep{zheng2024gpt}. More recent work removes these aids entirely, building agents that operate on raw screenshots and predict
  actions as pixel coordinates~\citep{qin2025ui, wang2025ui, andreux2025surfer, awadallah2025fara,gupta2026molmoweb,openaicomputeruse,geminicomputeruse}. We follow this direction and train visual web agents with screenshot input and coordinate-based actions.

\paragraph{Data and Environments for Web Agents.}

  Several efforts collect supervised fine-tuning (SFT) trajectories through human annotation or model-generated rollouts.                      
  Fara~\citep{awadallah2025fara} develops a multi-agent data generation system that produces 145K trajectories across 70K domains. MolmoWeb~\citep{gupta2026molmoweb} combines over 100K synthetic task trajectories with 30K+ human demonstrations and GUI perception data. OpenCUA~\citep{wang2025opencua} and AgentTrek~\citep{xu2024agenttrek} similarly collect demonstration data for web tasks. While valuable, SFT data alone provides limited support for the exploration and trial-and-error learning that RL training enables.

  Synthetic environments offer an alternative by enabling RL training in controlled settings. WebArena~\citep{zhou2023webarena} and
  VisualWebArena~\citep{koh2024visualwebarena} provide self-hosted websites that simulate e-commerce, forums, and content management systems.
  WebRL~\citep{qi2024webrl} and AgentGym-RL~\citep{xi2025agentgym} build on these for RL training, yet they cover only a handful of domains and do not
  capture the diversity of the real web.

  Recent work has explored scaling task generation. InstaV3~\citep{trabucco2025insta} develops an LLM-based pipeline to generate web navigation
  tasks across 146K live websites. WebGym~\citep{bai2026webgym} sources several datasets for RL training on live websites,
  but suffers from reproducibility issues and training instability due to timeouts and bot detection. Our framework addresses this through
  caching and LLM-based synthesis while remaining grounded in real websites and web navigation skills.

\paragraph{Evaluating Web Agents.}

  Evaluation benchmarks for web agents span visual grounding and end-to-end task completion.  For visual grounding, benchmarks like                                                                               
    ScreenSpot-v2~\citep{wu2024atlas}, ScreenSpot-Pro~\citep{li2025screenspot}, and MMBench-GUI~\citep{wang2025mmbench} evaluate an agent's ability to localize and interact with UI elements. For task completion, benchmarks vary in their use of simulated versus real environments. World of Bits~\citep{shi2017world} was an early effort that cached HTTP traffic to create reproducible offline approximations of websites, though limited to simple mini-tasks. WebArena~\citep{zhou2023webarena} and VisualWebArena~\citep{koh2024visualwebarena} evaluate agents on self-hosted websites with programmatic success checking. While reproducible, they suffer from a sim-to-real gap. Benchmarks on real websites include GAIA~\citep{mialon2023gaia}, WebVoyager~\citep{he2024webvoyager}, and Mind2Web~\citep{deng2023mind2web}, which test agents on live web tasks but face reproducibility
  challenges as websites change over time. WebVoyager additionally suffers from limited task diversity, with up to 51\% of tasks solvable via search shortcuts.                      
  Online-Mind2Web~\citep{xue2025illusion} addresses these issues with a more realistic setup that evaluates agents on live websites using an LLM-as-Judge for task success.  DeepShop~\citep{lyu2025deepshop} and WebTailBench~\citep{awadallah2025fara} further test agents on e-commerce and long-tail web tasks,            
  respectively. 

\begin{figure}[t]
  \centering
  \includegraphics[width=\linewidth]{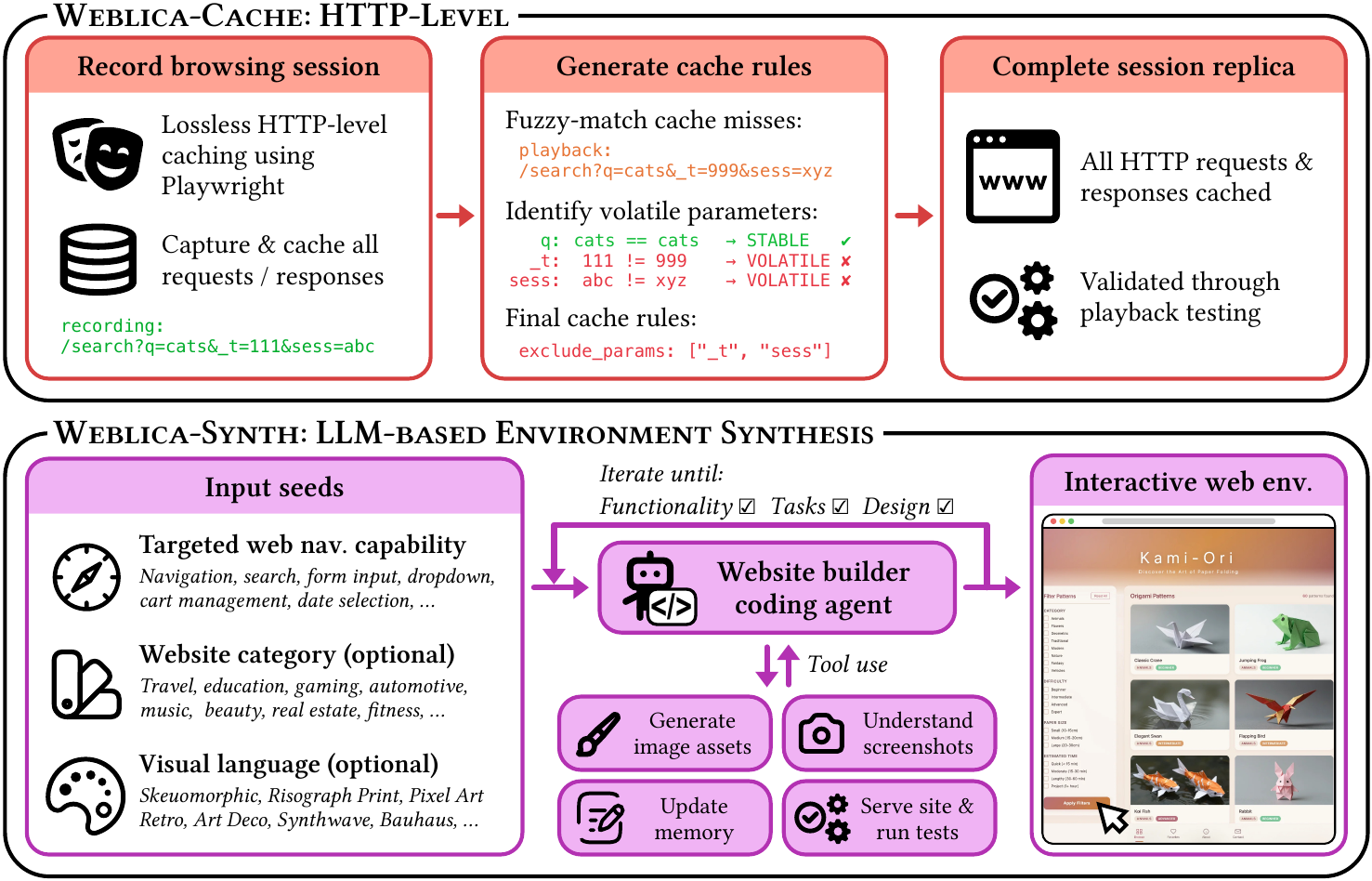}
  \vspace{-1em}
  
  \caption{Framework overview. \textbf{\methodcache (top)}: We record a browsing session capturing all HTTP
  traffic, then identify volatile parameters (e.g., timestamps, session tokens) that cause cache misses   
  during playback. These are used to generate site-specific caching rules that strip volatile parameters  
  from cache keys, enabling deterministic replay under complete network isolation. \textbf{\methodsynth      
  (bottom)}: We task coding agents to generate web environments parameterized by a navigation capability, 
  website category, and visual style. The agent writes framework-free HTML, CSS, and JavaScript and      
  iterates using tools (e.g., image generation, screenshot validation) until the website and tasks are
  functional. We apply both approaches at scale to create diverse offline training environments (see  
  \Cref{fig:pullfigure} for samples).}
  \label{fig:framework}
\end{figure}

\section{Framework}
\label{sec:framework}


\Cref{fig:framework} provides an overview of our framework. We first describe the agent formulation for visual web navigation (\Cref{sec:agent}), then detail the two environment  
  construction mechanisms: HTTP-level caching (\Cref{sec:caching}) and LLM-based synthesis (\Cref{sec:synthesis}).

\subsection{Agent Formulation}                                                                                                               
  \label{sec:agent}                                                                                                                                                                                               
\begin{figure}[t]
  \centering
  \includegraphics[width=\linewidth]{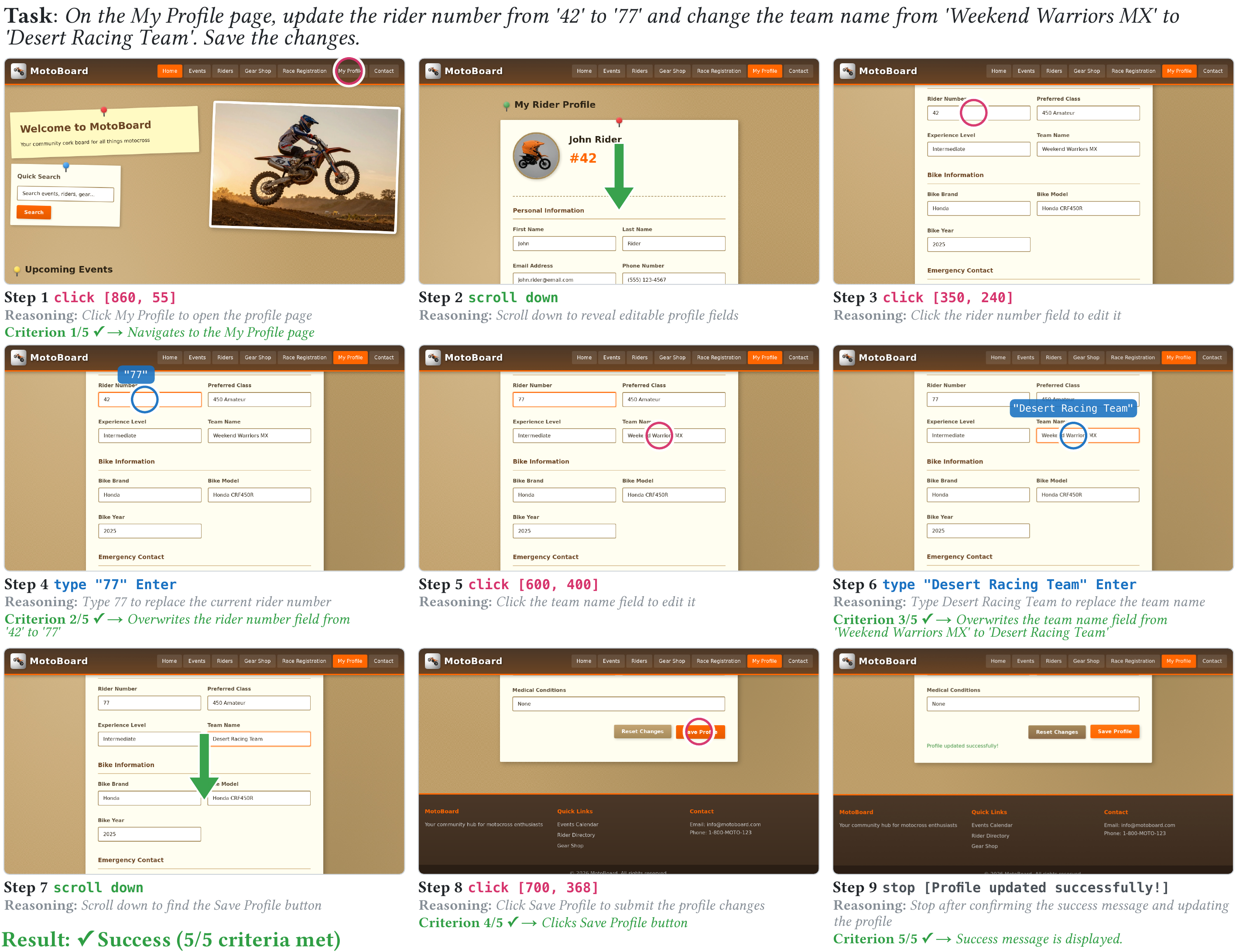}
  \caption{Example trajectory of \modelnameeightb solving a data entry task in a \methodsynth environment,
   evaluated by an LLM judge against task-specific criteria~(more examples in         
  \Cref{sec:appendix_additional_quals}).}
  \label{fig:trajectory_example}
  \end{figure}

  We formulate web navigation as a partially observable Markov decision process (POMDP) defined by the tuple $(\mathcal{S}, \mathcal{A},
  \mathcal{O}, T, R)$, where $\mathcal{S}$ is the state space of the browser, $\mathcal{A}$ is the action space, $\mathcal{O}$ is the          
  observation space, $T(s_{t+1} \mid s_t, a_t)$ is the transition function governing how the browser state changes in response to an action~(implemented via Playwright), and $R$ is the reward function (defined in \Cref{sec:reward}). At each timestep $t$, the agent receives an observation $o_t \in
  \mathcal{O}$ and selects an action $a_t \in \mathcal{A}$ conditioned on the task instruction $\tau$ and history $o_{\leq t}$. See \Cref{fig:trajectory_example} for an
   example trajectory.

  \paragraph{Observation Space.} Each observation $o_t = (s_t, u_t)$ consists of a browser screenshot $s_t$ rendered at $1280 \times 720$
  pixels and the current URL $u_t$. Unlike approaches that rely on accessibility trees or DOM structures, our agents operate purely on visual
  input.

   \paragraph{Action Space.} We adopt a coordinate-based action space following recent work on visual web agents. Coordinate-based actions (\texttt{click}, \texttt{hover}) take pixel
   positions $(x, y)$ as arguments, while other actions take task-specific arguments (text, keys, direction, etc.). The \texttt{stop} action terminates the episode and optionally 
  returns a response. Please see \Cref{tab:action_space} for the full action space.

  \paragraph{Policy.} The agent policy $\pi_\theta(a_t \mid o_{\leq t}, \tau)$ is parameterized by a vision-language model and follows a       
  ReAct-style~\citep{yao2022react} framework. At each step, the model produces a reasoning trace $r_t$ analyzing the current observation, 
  then selects an action $a_t$. Both reasoning traces and actions are appended to the history for subsequent steps. We use Qwen3-VL-Instruct as our base
   model, which supports coordinate-based action prediction without set-of-marks or other visual annotations.

\subsection{HTTP-Level Caching}
\label{sec:caching}

\textbf{Recording and replay.} We implement HTTP-level caching using Playwright to record and replay web interactions. During recording, we capture all HTTP traffic and index responses by normalized request    
  signatures. The key challenge is handling volatile parameters (timestamps, session tokens) that change between visits and cause cache misses. We address this with a rule-based
  normalization system that filters such parameters from URLs, headers, and POST bodies, with domain-specific rules and a multi-level fallback for progressive matching.

\textbf{Automated rule generation.} Developing caching rules for each website requires analyzing its traffic patterns. We automate this with a pipeline: first, we record a browsing session performed by a            
  Qwen3-VL-32B-Instruct agent, capturing all request parameters without filtering. A subsequent playback reveals cache misses, which we fuzzy-match against the recording to identify
   which parameters changed across visits. These reports are used to synthesize site-specific caching rules and synthetic responses for non-essential endpoints (e.g., analytics).   
  Generated rules are validated through playback with complete network isolation. Only sessions where the agent successfully completes the task under cached conditions are retained
   for training. This automated approach captures the full fidelity of real web content, including dynamic layouts and UI interactions, and scales to thousands of domains.

\textbf{Environments and tasks.} We leverage the InstaV3~\citep{trabucco2025insta} dataset as our task pool, which provides web navigation tasks across 146K websites generated  
  through an LLM-based pipeline. We match these tasks to cached environments and verify solvability under cached conditions, retaining 15.6K cached environments and tasks. We call  
  the resulting collection \methodcache.

\subsection{LLM-Based Environment Synthesis}
\label{sec:synthesis}

Recent agentic coding tools such as Claude Code~\citep{anthropic2025claudecode} have demonstrated strong autonomous coding capabilities. Given task descriptions and verification criteria, these systems can work independently until success requirements are satisfied. We leverage such tools to generate synthetic web environments at scale, enabling on-demand creation of functionality that is difficult to extract via caching alone, e.g. stateful tasks.

\textbf{Capability extraction.} We target the set of broad web navigation capabilities used across common websites and reflected in benchmarks like Online-Mind2Web~\citep{xue2025illusion}. To identify these capabilities, we first collect trajectories from Qwen3-VL-32B-Instruct attempting the Online-Mind2Web tasks. We then use GPT-5.2~\citep{openai2025gpt52} to analyze the successful and failed trajectory screenshots and extract coarse web interaction capabilities apparent in each. This yields 19,721 fine-grained capabilities (\textit{e.g., tab interface navigation, open dropdown menu, etc.}), which we aggregate into 144 higher-level capability groups (\textit{e.g., navigation, form input, date selection, map interaction}).

\textbf{Diverse website generation.} We use Claude Code (Opus~4.5~\citep{anthropic2025opus45}) to automatically generate self-contained web environments, writing static HTML, JavaScript, and CSS without any external framework dependencies. Each generation is parameterized by a target capability group, a website category randomly sampled from a pool of 1,160 domains (\textit{e.g., aviation, banking, yoga studio, zoology, etc.}), and a visual style sampled from 961 options (\textit{e.g., Editorial, Minimalist, Skeuomorphic, Duotone, etc.}). See \Cref{fig:app_om2w_distribution} for an overview of the distribution of capabilities, domains, and visual styles. Sampling website categories and visual styles is important for diversity, as we found that without them, generated websites converge to a narrow visual style and content. For each website we generate at least 10 tasks of varying difficulty that require performing the target capability and other interactions present on the site. The websites are static and have no backend, and we use JavaScript's \texttt{localStorage} feature to save values (e.g., items in cart) during a session. We use Z-Image-Turbo~\citep{Team2025ZImage} to automatically generate relevant visual assets for each website, such as product images, banners, etc. To ensure quality, we instruct Claude Code to self-validate each site by taking screenshots via Playwright and iterating until the output satisfies the desired quality, i.e., that the website and tasks are functional and that there are no CSS issues. 

\textbf{Environment server.} Serving synthetically generated websites instead of training on the real web also enables much faster environment interactions, with environment setup/reset and action-to-screenshot times around 50 to 150ms. Hosting environments locally removes the need to wait for the network to stabilize, and we additionally use Playwright's animation skipping feature. Together, these speed up Playwright by an order of magnitude (from $\sim$1.5s to 50-150ms per action), leading to an overall 30-40\% speed improvement in end-to-end RL training.

\textbf{Environments and tasks.} In total, we generated 310 sites covering the higher-level capability groups, and 2500 sites targeting the fine-grained capabilities sorted by frequency. We call the resulting collection of \textit{synthetic} web environments \methodsynth. From this full set, we reserve 2560 web environments for training covering 44,227 tasks (\method-train), and 250 sites for validation covering 500 tasks (\method-val). Please see \Cref{fig:pullfigure} as well as \Cref{fig:app_wvw_fractals_1,fig:app_wvw_fractals_2,fig:app_wvw_fractals_3} for random examples of synthetically generated websites.

\section{Training}
\label{sec:training}

We describe our training pipeline below. Please see \Cref{sec:appendix_training} for further details.

 \subsection{Data and Reward}\label{sec:reward}

  \paragraph{RL Data.} We describe task sourcing for RL environments in \Cref{sec:caching} and \Cref{sec:synthesis}. Please see \Cref{sec:appendix_data_stats} for          
  environment statistics.

  \paragraph{LLM-as-Judge Reward.} As many web navigation tasks are open-ended and cannot be evaluated programmatically or
  via string matching, we implement an LLM-as-judge~\citep{zheng2023judging} reward mechanism. Given a task description, the
  agent's action sequence, and resulting screenshots, we prompt GPT-4o~\citep{hurst2024gpt} to assess whether the agent
  successfully completed the task. This enables training on the full diversity of web tasks beyond those with programmatic
  verification. We validate the LLM judge by measuring agreement with human evaluations, finding 88\% agreement~(see \Cref{subsec:appendix_judge_prompt} for details).

  \paragraph{SFT Data.} We generate supervised fine-tuning data by collecting trajectories from a Qwen3-VL-32B-Instruct agent on InstaV3 queries. As not all rollouts succeed due to agent failures or website-specific issues such as changed task criteria or timeouts, we filter trajectories using the LLM judge to retain only successful completions. To increase coverage, we sample multiple rollouts per query with diverse sampling parameters, yielding 51.7K SFT trajectories in total. Please see \Cref{fig:sft_traj_hist} for the trajectory
   length distribution.

\subsection{Post-Training Stages}

\paragraph{SFT Warm-Start.} We fine-tune the base VLM on verified trajectories before RL training. This provides a strong  
  initialization that already exhibits reasonable web navigation behavior. We ablate the role of SFT by comparing
  warm-started RL against cold-start RL that trains directly from the base model in \Cref{sec:ablations}.

\paragraph{RL Training.} We train using Dr.~GRPO~\citep{liu2025understanding}, a variant of Group Relative Policy Optimization (GRPO)~\citep{shao2024deepseekmath}, with the         
  LLM-as-Judge reward on our diverse environment suite. During RL, we oversample rollouts and filter to retain groups with mixed success signals, similar to
  DAPO~\citep{yu2025dapo}, and apply fallback mechanisms to handle errors or timeouts, ensuring stable training.

\section{Experiments}
\label{sec:experiments}

We evaluate \method on web navigation and UI grounding benchmarks. After describing our setup (\Cref{sec:setup}), we present main results against existing agents              
  (\Cref{sec:main_results}), analyze test-time scaling (\Cref{sec:test_time_scaling}), and ablate training stages, environment choices, and the effect on grounding          
  (\Cref{sec:ablations}). 
  
\subsection{Setup}\label{sec:setup}

\paragraph{Environments.} Our training environment consists of \methodcache and \methodsynth environments, covering diverse domains including e-commerce, information,             
  entertainment, and government services. We subsample 10K tasks from each, yielding 20K tasks for RL training. 
  
\begin{table}[t]                                                                                                                                                                   
    \caption{Comparison across web navigation benchmarks. Avg.\ is the mean over the three live                                                                                        
  benchmarks. Baseline results are reported from~\citep{awadallah2025fara,gupta2026molmoweb}. Italic Online-Mind2Web results are from the auto-eval leaderboard.                                                                                             
  With only 60 total steps compared to $\geq$100 for baselines, \modelnameeightb already                                                                                             
  achieves the best average among non-proprietary models (above dashed line), and further                                                                                            
  improves with additional test-time compute (below dashed line). We \textbf{bold} the best                                                                                          
  non-proprietary result per column and \underline{underline} the second best.                                                                                                       
  $^\dagger$ Fara-7B and MolmoWeb-8B allow up to 5 and 10 retries per trajectory.}                                                                                                         
    \label{tab:main_results}                                                                                                                                                         
    \centering                                                                                                                                                                       
    \resizebox{\textwidth}{!}{                                                                                                                                                       
    \begin{tabular}{@{}lc ccc >{\columncolor{gray!8}}c @{\hskip 0.8em} c@{}}                                                                                                         
      \toprule                                                                                                                                                                       
      & & \multicolumn{4}{c}{\textit{Live Web Navigation}} & \textit{Synthetic} \\                                                                                                   
      \cmidrule(lr){3-6} \cmidrule(l){7-7}                                                                                                                                           
      Model & Total \# Steps & Online-Mind2Web & DeepShop & WebTailBench & Avg. &                                                                                                    
      \method-val \\                                                                                                                                                                 
      \midrule                                                                                                                                                                       
      \multicolumn{7}{l}{\textbf{API-only}} \\                                                                                                                                       
      OpenAI CUA & 100 & \textit{58.3} & 24.7 & 25.7 & 36.2 & -- \\                                                                                                                 
      Gemini CUA & 100 & \textit{57.3} & 62.0 & 63.0 & 60.8 & -- \\                                                                                                                  
      Yutori Navigator~\citep{yutori2025navigator} & -- & \textit{64.7} & -- & -- & -- & -- \\                                                                                        
      \midrule                                                                                                                                                                       
      \multicolumn{7}{l}{\textbf{Open-weight}} \\                                                                                                                                    
      Qwen3-VL-Instruct-8B & 30 & 28.6 & 24.1 & 21.8 & 24.8 & 56.9 \\                                                                                                                
      UI-TARS-1.5-7B~\citep{qin2025ui} & 100 & 31.3 & 11.6 & 19.5 & 20.8 & -- \\                                                                                                     
      GLM-4.1V-9B-Thinking~\citep{hong2025glm} & 100 & 33.9 & 32.0 & 22.4 & 29.4 & -- \\                                                                                             
      Fara-7B~\citep{awadallah2025fara} & $\geq$100$^\dagger$ & 34.1 & 26.2 & 38.4 & 32.9 & -- \\                                                                                    
      MolmoWeb-8B~\citep{gupta2026molmoweb} & $\geq$100$^\dagger$ & 35.3 & \underline{42.3} & \textbf{49.5} & \underline{42.4} & -- \\                                                
      \midrule                                                                                                                                                                       
      \multicolumn{7}{l}{\textbf{Ours (pass@k)}} \\                                                                                                                                  
      \modelnameeightb (k=1) & 30 & \underline{39.2} & 34.2 & 33.5 & 35.6 & \underline{70.6} \\                                                                                      
      \modelnameeightb (k=2) & \textbf{60} & \textbf{50.3} & \textbf{45.4} & \underline{47.0} & \textbf{47.6} & \textbf{79.0} \\                                                     
      \cdashline{1-7}                                                                                                                                                                     
      \noalign{\vskip 2pt}                                                                                                                                                           
      \modelnameeightb (k=4) & 120 & 60.5 & 55.9 & 60.3 & 58.9 & 84.7 \\                                                                                                             
      \modelnameeightb (k=8) & 240 & 68.8 & 65.8 & 72.2 & 68.9 & 88.6 \\                                                                                                             
      \bottomrule                                                                                                                                                                    
    \end{tabular}                                                                                                                                                                    
    }                                                                                                                                                                                
  \end{table} 

  \paragraph{Evaluation.} We evaluate on four web navigation benchmarks. Online-Mind2Web (OM2W)~\citep{xue2025illusion} tests agents on diverse live websites.
  DeepShop~\citep{lyu2025deepshop} focuses on e-commerce tasks. WebTailBench (WTB)~\citep{awadallah2025fara} tests agents on long-tail web tasks. We use the official judges provided
  by each benchmark for evaluation. For WebTailBench, we follow MolmoWeb~\citep{gupta2026molmoweb} and use the WebVoyager~\citep{he2024webvoyager} judge. We also evaluate on \method-val, 500 held-out tasks on 250 unseen synthesized environments, to measure in-distribution generalization. We study test-time scaling by running $k \in \{1,
   2, 4, 8\}$ independent attempts and reporting pass@$k$, where total step budget scales linearly with $k$. Unlike retry-based evaluation in prior work~\citep{awadallah2025fara,    
  gupta2026molmoweb}, where failures attributed to the environment are selectively discarded, this provides a more reliable estimate of task success and a transparent measure of
  total test-time compute. We repeat each experiment three times and report mean accuracy; standard deviations are reported in plots and in \Cref{sec:appendix_additional_quants}. We
  additionally evaluate UI grounding on MMBench-GUI~\citep{wang2025mmbench}, ScreenSpot-v2~\citep{wu2024atlas}, and ScreenSpot-Pro~\citep{li2025screenspot} (\Cref{sec:grounding}).

  \paragraph{Baselines.} We compare against open-weight models including UI-TARS-1.5-7B~\citep{qin2025ui}, GLM-4.1V-9B-Thinking~\citep{hong2025glm}, Fara-7B~\citep{awadallah2025fara}, 
  and MolmoWeb-8B~\citep{gupta2026molmoweb}, as well as API-based agents including OpenAI computer-use-preview, Gemini computer-use-preview, and Yutori
  Navigator~\citep{yutori2025navigator}. We also report the base Qwen3-VL-Instruct-8B as a reference.

\subsection{Main Results}\label{sec:main_results}

\Cref{tab:main_results} presents results across four benchmarks and \Cref{fig:trajectory_example} shows a qualitative example (more in \Cref{sec:appendix_additional_quals}). We summarize the key observations below.
\paragraph{Training on \method data improves the base model by 44\%.} At pass@1 with 30 steps, \modelnameeightb improves the base Qwen3-VL-8B from 24.8\% to 35.6\% average across 
  the three live benchmarks, an absolute gain of 10.8 percentage points. On \method-val, our held-out evaluation set of synthesized environments, training improves the base from 56.9\% 
  to 70.6\%. We observe consistent improvements at 2B and 4B model sizes as well (\Cref{fig:pullfigure} and \Cref{subsec:appendix_model_sizes}).
  
  \paragraph{\modelnameeightb outperforms open-weight baselines with less test-time compute.} With only 60 total steps, \modelnameeightb surpasses the best open-weight      
  baseline average (47.6\% vs 42.4\% for MolmoWeb-8B at $\geq$100 steps), and scales to 58.9\% and 68.9\% with 120 and 240 steps respectively.                                                                    
  \paragraph{\modelnameeightb is competitive with API models.} At pass@1 with only 30 steps, \modelnameeightb already matches OpenAI computer-use-preview (35.6\% vs 36.2\% at 100   
  steps). Additional test-time compute leads to further improvements: with 120 total steps, \modelnameeightb   
  reaches 58.9\%, approaching Gemini computer-use-preview (60.8\%), and with 240 steps reaches 68.9\%. These results demonstrate the effectiveness of training on synthetic
  \method environments, producing strong visual web agents without relying on large-scale human demonstrations.

\subsection{Test-Time Scaling}
  \label{sec:test_time_scaling}

\begin{figure}[t]                                                                                                                                                                  
    \centering                                                                                                                                                                     
    \includegraphics[width=\linewidth]{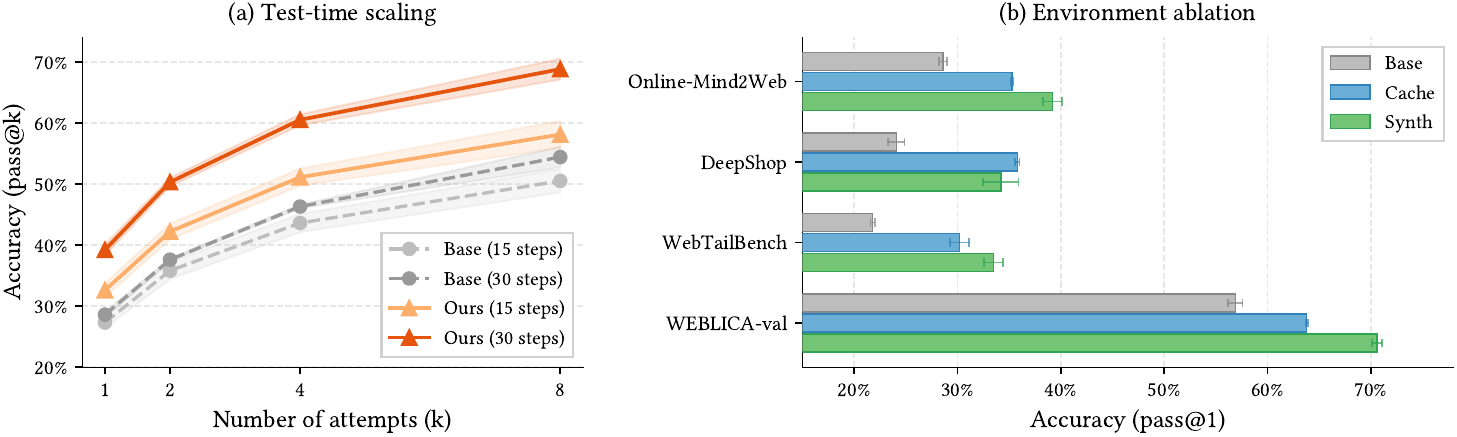}                                                                                                             
  \caption{\textbf{(a)} Test-time scaling on Online-Mind2Web. \modelnameeightb benefits from both increasing the action budget (steps per episode) and parallel attempts (pass@$k$).   
  The base Qwen3-VL-8B, in contrast, shows minimal gains from longer episodes, suggesting that RL training enables the model to effectively leverage additional steps. \textbf{(b)}  
  Environment ablation across four benchmarks. Both training configurations improve substantially over the base model, with synthesized environments outperforming cache-only on most
   benchmarks.}
    \label{fig:scaling_env}                                                          
  \end{figure}

  We study how test-time compute scaling affects performance through two axes: the number of parallel attempts (pass@$k$) and the per-episode action budget.                         
  \Cref{fig:scaling_env}a compares \modelnameeightb and the base Qwen3-VL-8B on Online-Mind2Web. \modelnameeightb improves consistently along both axes: increasing the action budget from 
  15 to 30 steps improves pass@1 from 32.6\% to 39.2\%, and pass@8 at 30 steps per attempt reaches 68.8\%. In contrast, the base model shows minimal gains from a larger action budget, with its 
  15- and 30-step curves nearly overlapping. This gap suggests that RL training is key to unlocking effective use of longer episodes.

\subsection{Analysis}\label{sec:ablations}

\paragraph{Environment Type.}\label{sec:env_ablation}

We compare the effect of training with \methodcache only versus \methodsynth only. \Cref{fig:scaling_env}b shows pass@1 results across   
  all four benchmarks. Both configurations improve substantially over the base model. Synthesized environments outperform cache-only on most benchmarks, with the largest gains on   
  Online-Mind2Web (39.2\% vs 35.3\%) and WebTailBench (33.5\% vs 30.2\%), while cache-only performs slightly better on DeepShop (35.8\% vs 34.2\%). Our initial attempts with        
  different mixture ratios of cached and synthesized environments did not yield further improvements; we expect that designing effective training curricula across environment types
  could lead to additional gains.

\paragraph{Training Stages and Model Size.}\label{sec:train_stages}

\begin{figure}[t]
      \centering
      \includegraphics[width=\linewidth]{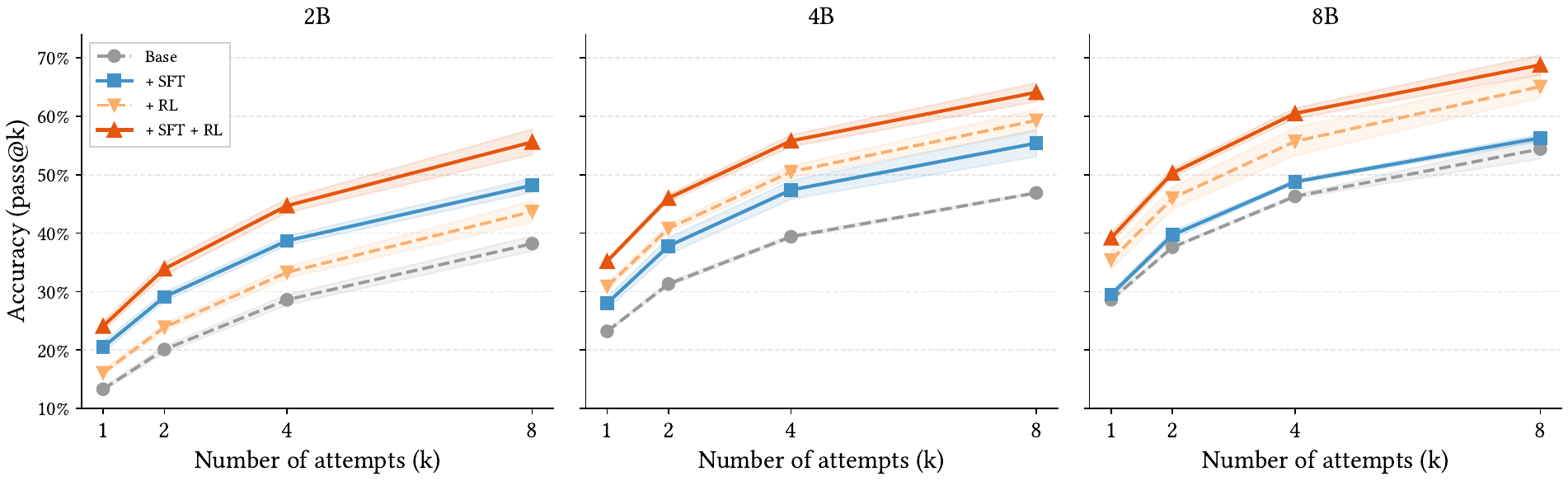}
      \caption{Training stage ablation on Online-Mind2Web across three model sizes. Both SFT and RL individually improve over the base model, but their combination performs best at all 
  scales.}
      \label{fig:training_ablation}
  \end{figure}

\Cref{fig:training_ablation} compares the contribution of each training stage across model sizes on Online-Mind2Web. Both SFT and RL provide consistent gains at all scales. 
  At 2B, the full SFT+RL pipeline improves pass@1 from 13.3\% to 24.1\%, at 4B from 23.2\% to 35.2\%, and at 8B from 28.6\% to 39.2\%. RL on top of SFT provides a larger improvement
   than SFT alone at every scale. Comparing SFT$\to$RL with Base$\to$RL shows that SFT initialization is critical for smaller models, but SFT data has diminishing returns at larger 
  scales.

\begin{table}[h]                                                                                                                                                                   
    \centering
    \caption{\modelnameeightb preserves grounding despite no grounding-specific training data.}                                 
    \label{tab:grounding}                                                                                                                                                            
    \footnotesize                                                                                                                                                                    
    \begin{tabular}{@{}lccc@{}}                                                                                                                                                      
      \toprule                                                                                                                                                                       
      Model & MMBench-GUI & ScreenSpot-v2 & ScreenSpot-Pro \\                                                                                                                        
      \midrule                                                                                                                                                                       
      Fara-7B & --- & 89.3 & --- \\                                                                                                                                                  
      MolmoWeb-Ground-8B & --- & 91.8 & --- \\                                                                                                                                       
      \midrule                                                                                                                                                                       
      Qwen3-VL-Instruct-8B & 82.85 & 93.95 & 54.71 \\                                                                                                                                
      \modelnameeightb & \textbf{83.74} & \textbf{94.50} & \textbf{55.28} \\                                                                                                         
      \bottomrule                                                                                                                                                                    
    \end{tabular}                                                                                                                                                                    
  \end{table}

\paragraph{Effect on Grounding.}\label{sec:grounding}
      
  As visual grounding is a core capability for web agents, we evaluate whether our training pipeline affects it, given that our training data contains no grounding-specific data.   
  \Cref{tab:grounding} shows that grounding is preserved after training, with modest improvements across all three benchmarks. Since grounding performance remains comparable,  
  the gains on web navigation benchmarks stem from improved navigation behavior rather than better visual grounding.

\section{Conclusion and Limitations}
\label{sec:conclusion}

We presented \method, a framework for building scalable and reproducible training environments for visual web agents. \method combines HTTP-level caching of real websites with
   LLM-based synthesis of interactive environments, enabling large-scale RL training without the instability of live web interaction. Our best model, \modelnameeightb,  
  achieves strong results across several web navigation benchmarks compared to open-weight baselines of similar size, and scales favorably with additional test-time compute. We outline some limitations of our work and future directions.
  \begin{itemize}
      \item \textbf{Cached environments} provide a narrow and partial view of live websites. They are static snapshots that do not reflect updates over time,
  and do not capture the full complexity of dynamic web applications. While we observe clear gains from training on cached environments, further exploration of
   methods to close this gap is an interesting direction.

      \item \textbf{Synthesized environments} capture core navigation patterns but do not yet model all aspects of real websites. A sim-to-real gap remains, which could be further closed with stronger generative models that produce more faithful and diverse website designs.

      \item \textbf{Single-turn tasks.} Our current setup evaluates agents on isolated tasks with a fixed goal defined at the start of the episode. Real-world web usage involves multi-turn sessions with evolving goals, human-in-the-loop interaction where the user provides feedback or corrections mid-session, and personalization aspects such as
  adapting to user preferences and maintaining memory across sessions. Extending to these settings is an exciting direction.

      \item \textbf{Training.} We obtain promising gains using a vanilla RL training formulation. Further exploration of the RL framework, including
  long-horizon RL and scaling RL compute~\citep{khatri2025art}, and richer SFT data such as error recovery trajectories are interesting directions for improving web agent training.

      \item \textbf{Beyond web.} Our framework currently targets web navigation. Extending it to other GUI environments such as mobile and desktop applications to build generalist computer-use agents is a promising direction.
  \end{itemize}

\section*{Acknowledgements}                                                           
  We thank Jesse Allardice, Mingfei Gao, Rui Tian, and Ege Özsoy for their help with the project, and Andrew Szot, Alexander Toshev, and Kaixin Ma for feedback and discussions.

\applefootnote{ \textcolor{textgray}{\sffamily Apple and the Apple logo are trademarks of Apple Inc., registered in the U.S. and other countries and regions.}}

\clearpage
\newpage

\bibliographystyle{plainnat}                                                                                     \bibliography{main}


\newpage

\newpage                                                                                                                                                                           
\appendix                                                   
\crefalias{section}{appsec}
\crefalias{subsection}{appsec}
\crefalias{subsubsection}{appsec}
\section*{\LARGE Appendix}                                                                                 
\addcontentsline{toc}{section}{Appendix}                                                                     
\startcontents[appendix]                                                                      
\printcontents[appendix]{l}{1}{\setcounter{tocdepth}{2}}                                  

\newpage

\section{Additional Quantitative Results}\label{sec:appendix_additional_quants}

\subsection{Results across model sizes}\label{subsec:appendix_model_sizes}                                                                                                                
\begin{figure}[h]                                                                                                                                                                  
    \centering                                                                                                                                                               
    \includegraphics[width=\linewidth]{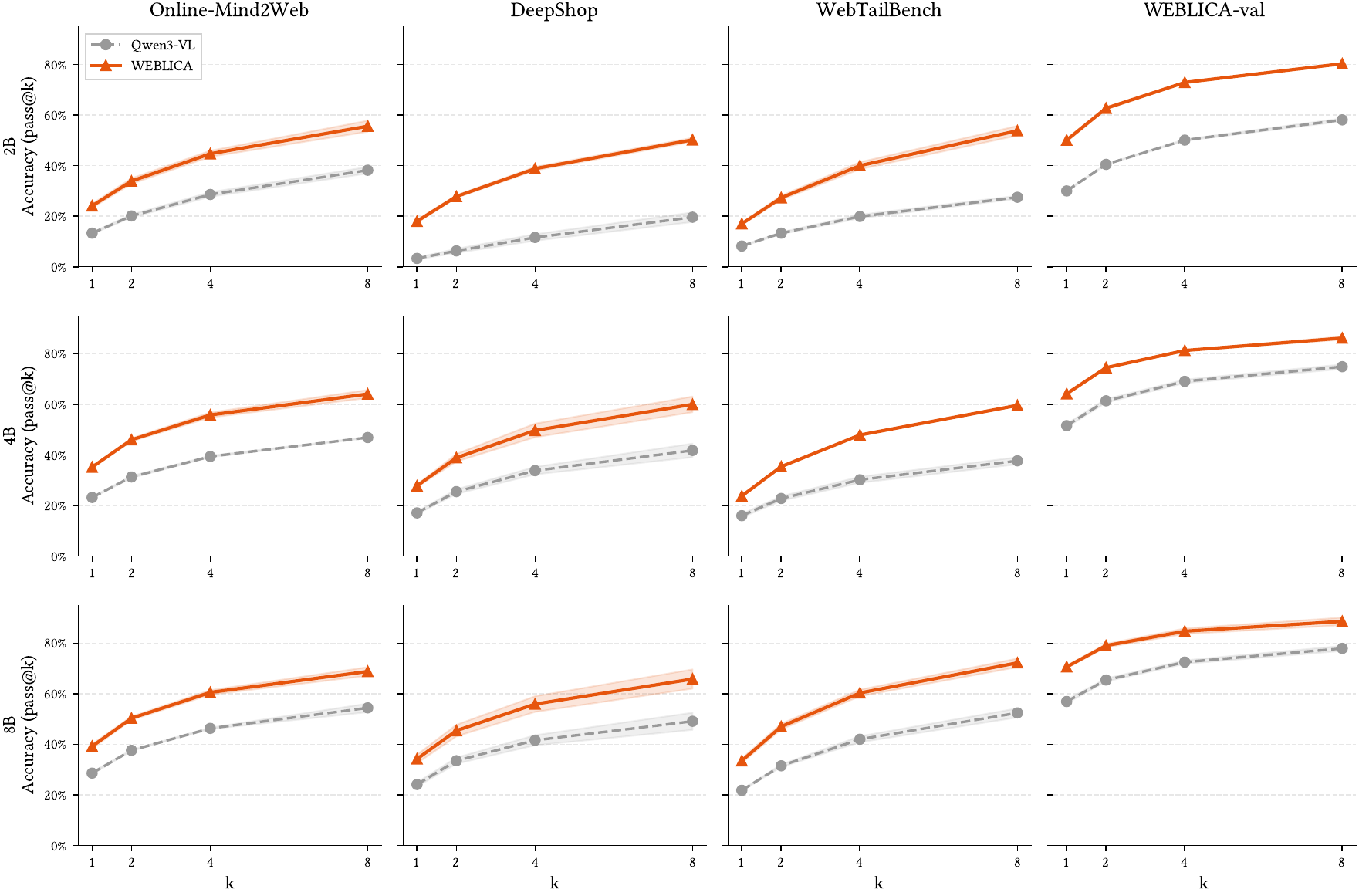}                                                                                                             
    \caption{Pass@k accuracy for \modelnametwob, \modelnamefourb, and \modelnameeightb (orange) compared to their respective Qwen3-VL-Instruct base models (gray) across four          
  benchmarks.}                             
    \label{fig:appendix_pass_k}                                                                                                                                                      
  \end{figure}

  \Cref{fig:appendix_pass_k} shows pass@k results across all three model sizes and four benchmarks, with corresponding numerical values in \Cref{tab:appendix_pass_k}.
  Training consistently improves over the base model at all scales and evaluation points.                                                                                                         

\begin{table}[h]                                                                                                                                                                   
    \caption{Pass@k accuracy across model sizes and benchmarks.}                                                                                        
    \label{tab:appendix_pass_k}                                                                                                                                                      
    \centering                                                                                                                                                                       
    \begin{tabular}{@{}ll r@{$\,\pm\,$}l r@{$\,\pm\,$}l r@{$\,\pm\,$}l r@{$\,\pm\,$}l @{}}                                                                                           
      \toprule                                                                                                                                                                       
      & & \multicolumn{2}{c}{pass@1} & \multicolumn{2}{c}{pass@2} & \multicolumn{2}{c}{pass@4} & \multicolumn{2}{c}{pass@8} \\                                                       
      \midrule                                                                                                                                                                       
      \multicolumn{10}{l}{\textbf{Model size: 2B}} \\                                                                                                                                
      \addlinespace[2pt]                                                                                                                                                             
      \multirow{2}{*}{Online-Mind2Web} & Qwen3-VL & 13.3 & 0.3 & 20.1 & 0.6 & 28.6 & 0.9 & 38.2 & 1.3 \\                                                                             
      & \modelnametwob & 24.1 & 0.8 & 33.9 & 1.1 & 44.7 & 1.2 & 55.6 & 2.2 \\                                                                                                        
      \addlinespace[2pt]                                                                                                                                                             
      \multirow{2}{*}{DeepShop} & Qwen3-VL & 3.3 & 0.4 & 6.3 & 0.8 & 11.6 & 1.3 & 19.6 & 1.9 \\                                                                                      
      & \modelnametwob & 18.0 & 0.6 & 27.8 & 0.5 & 38.8 & 0.7 & 50.2 & 0.8 \\                                                                                                        
      \addlinespace[2pt]                                                                                                                                                             
      \multirow{2}{*}{WebTailBench} & Qwen3-VL & 8.2 & 0.2 & 13.3 & 0.4 & 19.9 & 0.7 & 27.5 & 0.6 \\                                                                                 
      & \modelnametwob & 17.0 & 0.4 & 27.3 & 0.8 & 40.0 & 1.4 & 53.8 & 1.9 \\                                                                                                        
      \addlinespace[2pt]                                                                                                                                                             
      \multirow{2}{*}{\method-val} & Qwen3-VL & 30.0 & 0.3 & 40.5 & 0.1 & 50.1 & 0.3 & 58.1 & 0.6 \\                                                                                 
      & \modelnametwob & 50.1 & 0.6 & 62.7 & 0.5 & 72.9 & 0.2 & 80.3 & 0.2 \\                                                                                                        
      \midrule                                                                                                                                                                       
      \multicolumn{10}{l}{\textbf{Model size: 4B}} \\                                                                                                                                
      \addlinespace[2pt]                                                                                                                                                             
      \multirow{2}{*}{Online-Mind2Web} & Qwen3-VL & 23.2 & 0.2 & 31.3 & 0.4 & 39.4 & 0.3 & 46.9 & 0.2 \\                                                                             
      & \modelnamefourb & 35.2 & 0.3 & 46.0 & 0.6 & 55.8 & 1.0 & 64.1 & 1.6 \\                                                                                                       
      \addlinespace[2pt]                                                                                                                                                             
      \multirow{2}{*}{DeepShop} & Qwen3-VL & 17.1 & 0.5 & 25.5 & 0.9 & 33.8 & 1.5 & 41.8 & 2.7 \\                                                                                    
      & \modelnamefourb & 27.8 & 0.5 & 38.9 & 1.5 & 49.7 & 2.7 & 60.0 & 3.1 \\                                                                                                       
      \addlinespace[2pt]                                                                                                                                                             
      \multirow{2}{*}{WebTailBench} & Qwen3-VL & 16.0 & 0.7 & 22.8 & 0.9 & 30.2 & 1.2 & 37.7 & 1.4 \\                                                                                
      & \modelnamefourb & 23.8 & 0.4 & 35.4 & 0.3 & 47.9 & 0.3 & 59.6 & 0.3 \\                                                                                                       
      \addlinespace[2pt]                                                                                                                                                             
      \multirow{2}{*}{\method-val} & Qwen3-VL & 51.6 & 0.7 & 61.4 & 0.7 & 69.1 & 0.7 & 74.9 & 0.9 \\                                                                                 
      & \modelnamefourb & 64.2 & 0.5 & 74.5 & 0.3 & 81.3 & 0.2 & 86.2 & 0.6 \\                                                                                                       
      \midrule                                                                                                                                                                       
      \multicolumn{10}{l}{\textbf{Model size: 8B}} \\                                                                                                                                
      \addlinespace[2pt]                                                                                                                                                             
      \multirow{2}{*}{Online-Mind2Web} & Qwen3-VL & 28.6 & 0.4 & 37.6 & 0.2 & 46.3 & 0.3 & 54.4 & 1.7 \\                                                                             
      & \modelnameeightb & 39.2 & 0.9 & 50.3 & 0.7 & 60.5 & 0.9 & 68.8 & 1.7 \\                                                                                                      
      \addlinespace[2pt]                                                                                                                                                             
      \multirow{2}{*}{DeepShop} & Qwen3-VL & 24.1 & 0.8 & 33.5 & 1.2 & 41.6 & 2.0 & 49.1 & 3.4 \\                                                                                    
      & \modelnameeightb & 34.2 & 1.7 & 45.4 & 2.3 & 55.9 & 3.1 & 65.8 & 3.8 \\                                                                                                      
      \addlinespace[2pt]                                                                                                                                                             
      \multirow{2}{*}{WebTailBench} & Qwen3-VL & 21.8 & 0.2 & 31.5 & 0.5 & 42.0 & 1.2 & 52.4 & 1.8 \\                                                                                
      & \modelnameeightb & 33.5 & 0.9 & 47.0 & 1.1 & 60.3 & 1.3 & 72.2 & 1.7 \\                                                                                                      
      \addlinespace[2pt]                                                                                                                                                             
      \multirow{2}{*}{\method-val} & Qwen3-VL & 56.9 & 0.7 & 65.4 & 0.6 & 72.5 & 0.7 & 77.9 & 1.0 \\                                                                                 
      & \modelnameeightb & 70.6 & 0.5 & 79.0 & 0.7 & 84.7 & 1.1 & 88.6 & 1.5 \\                                                                                                      
      \bottomrule                                                                                                                                                                    
    \end{tabular}                                                                                                                                                                    
  \end{table}

\subsection{Downstream performance during training} \label{subsec:appendix_training_curve}

\begin{figure}[h]                                                                                                                                                                  
    \centering                                                                       
    \includegraphics[width=0.85\linewidth]{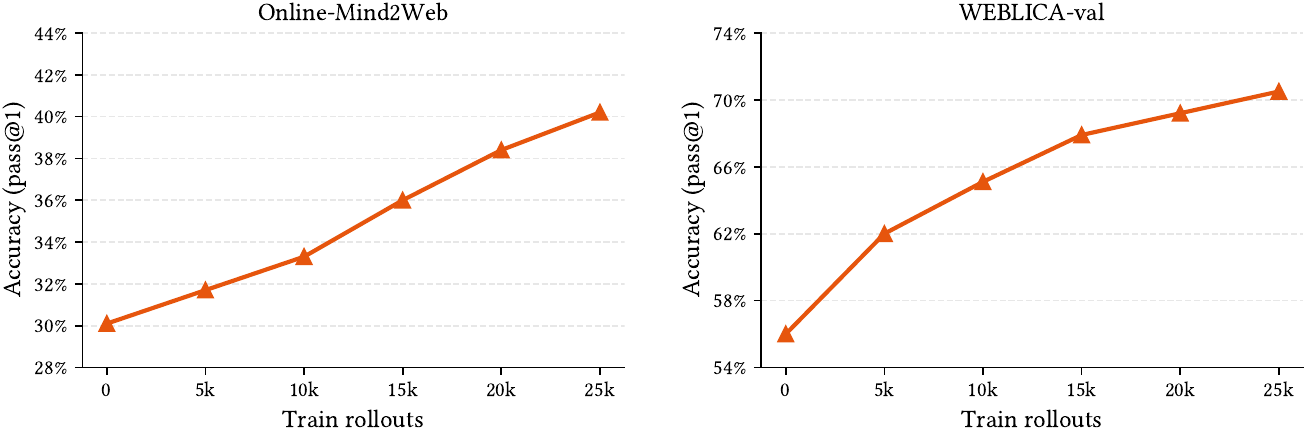}                                                                                                               
    \caption{Pass@1 accuracy on Online-Mind2Web (left) and \method-val (right) during RL training for \modelnameeightb. Each training step uses 1024 rollouts. The first data point  
  (0 rollouts) is the SFT-initialized model before RL.}                                                                                                                              
    \label{fig:training_curve}                                                                                                                                                       
  \end{figure}

\Cref{fig:training_curve} shows steady improvement in pass@1 accuracy on Online-Mind2Web and \method-val throughout RL training for \modelnameeightb for a sample training run.                      

\section{Additional Qualitative Results}\label{sec:appendix_additional_quals}

\subsection{\methodsynth visualizations}
\begin{figure}[H]
  \centering
  \includegraphics[width=0.85\linewidth]{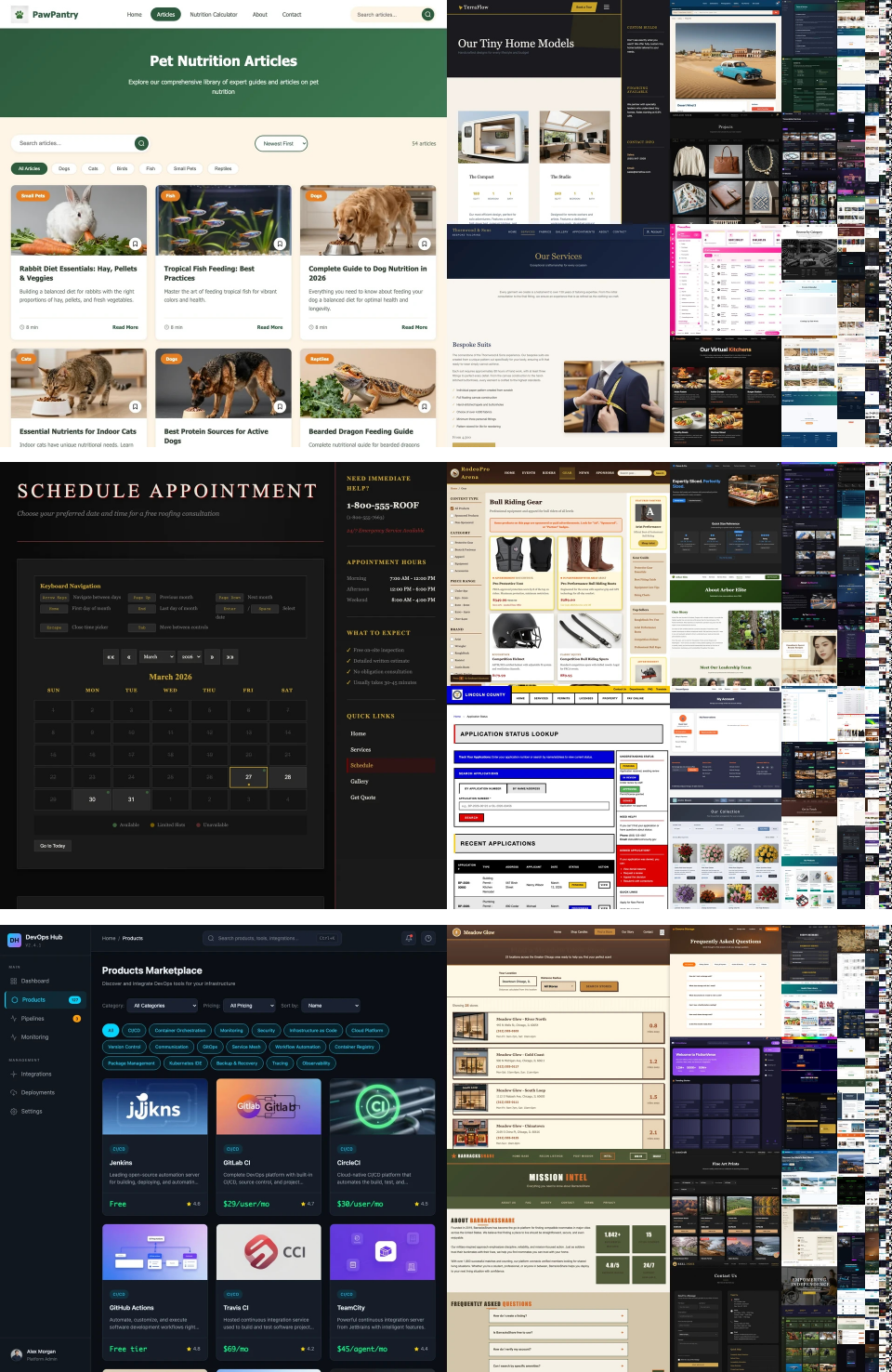}
  \caption{\textbf{\methodsynth samples}: Additional visualizations of synthetically generated web environments, capturing a broad set of web capabilities.}
  \label{fig:app_wvw_fractals_1}
\end{figure}

\begin{figure}[H]
  \centering
  \includegraphics[width=0.85\linewidth]{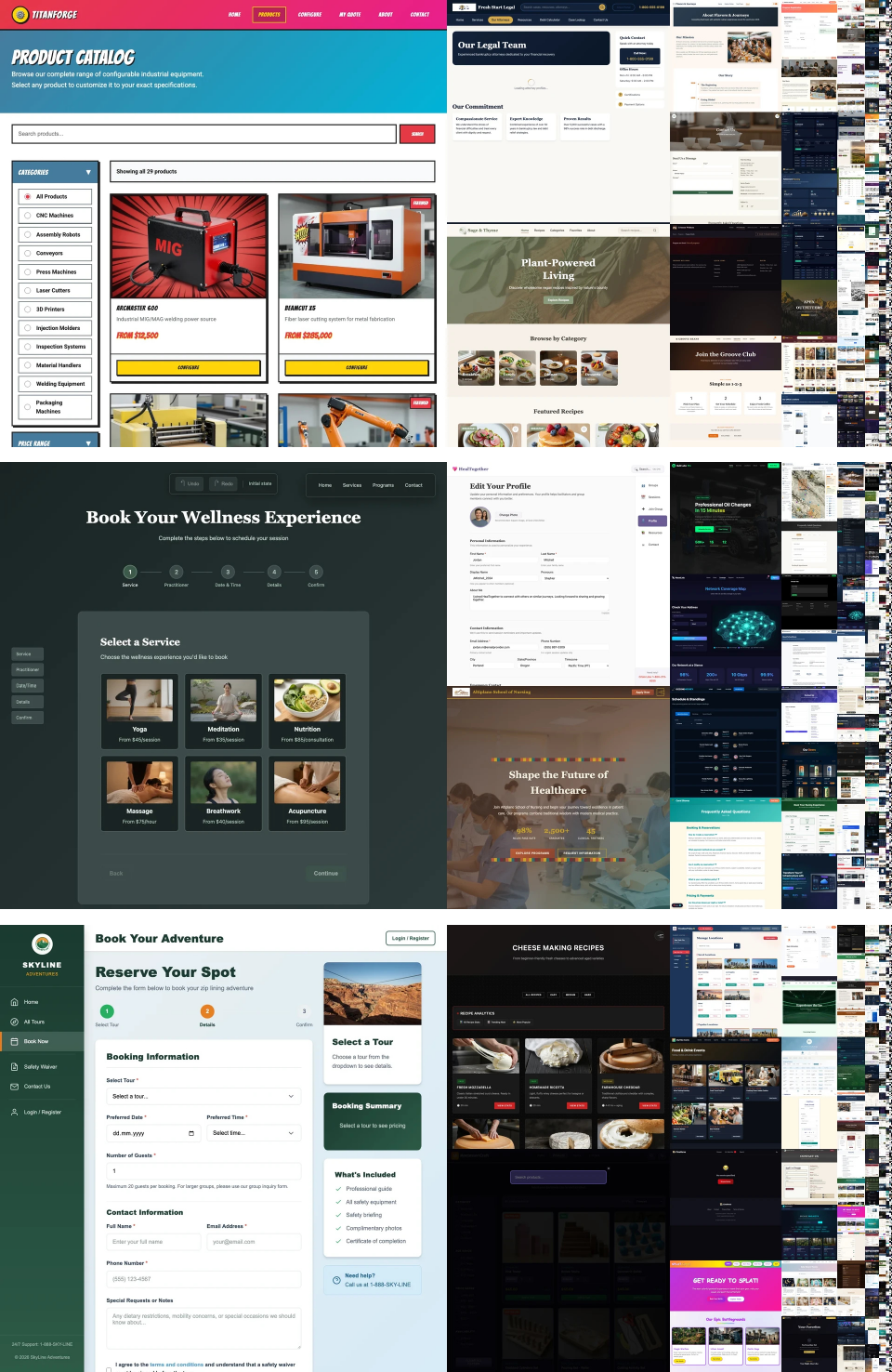}
  \caption{\textbf{\methodsynth samples}: \Cref{fig:app_wvw_fractals_1} cont.}
  \label{fig:app_wvw_fractals_2}
\end{figure}

\begin{figure}[H]
  \centering
  \includegraphics[width=0.85\linewidth]{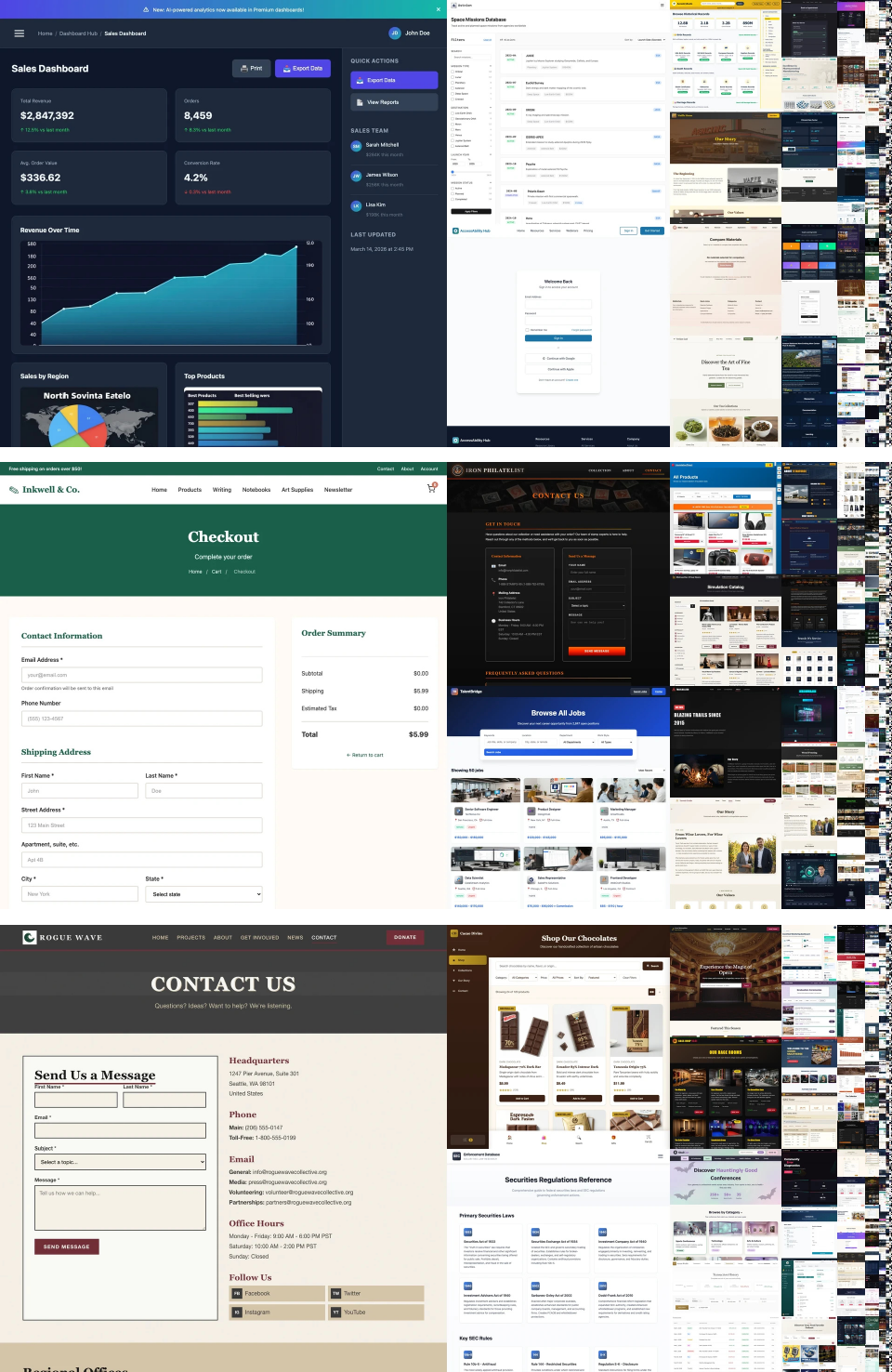}
  \caption{\textbf{\methodsynth samples}: \Cref{fig:app_wvw_fractals_1} and \Cref{fig:app_wvw_fractals_2} cont.}
  \label{fig:app_wvw_fractals_3}
\end{figure}

We provide additional random sample screenshots of \methodsynth synthetically generated web environments in \Cref{fig:app_wvw_fractals_1,fig:app_wvw_fractals_2,fig:app_wvw_fractals_3}. The generated pages target a broad distribution of web capabilities and are visually diverse.

\newpage 
\clearpage

\newpage
\subsection{Solved trajectory examples}\label{appendix:sec_solved_trajectories}
\begin{figure}[H]
  \centering
  \includegraphics[width=\linewidth]{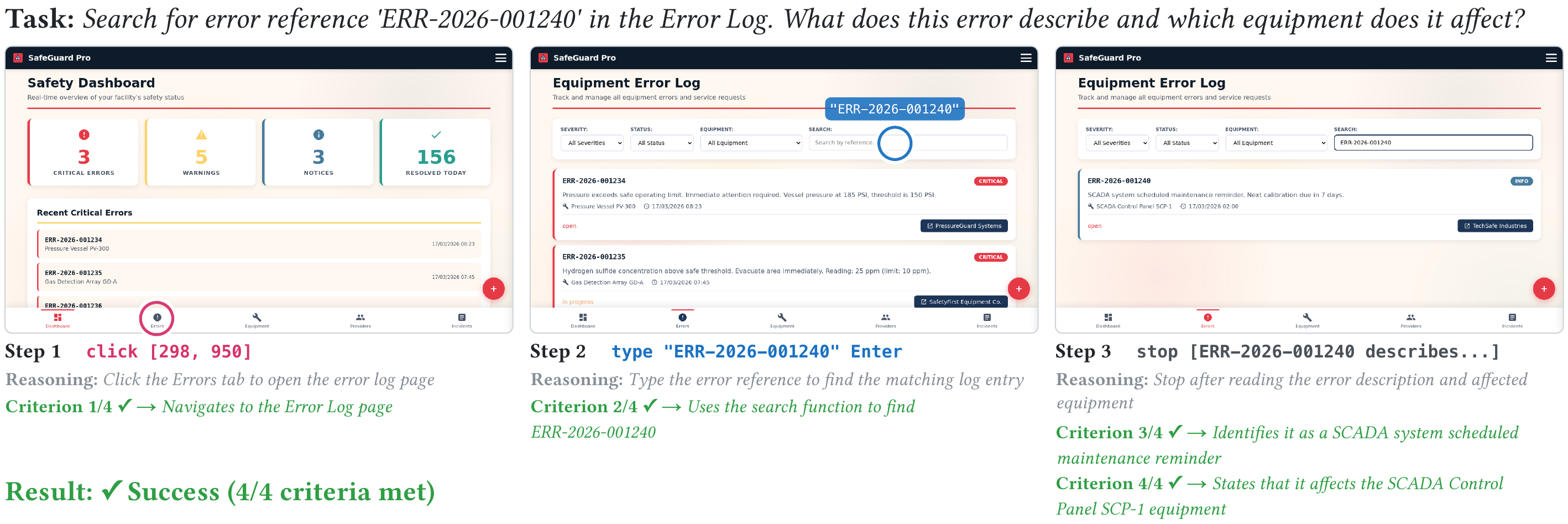}
  \caption{Example trajectory of \modelnameeightb solving a \method-val task.}
  \label{fig:app_trajectory_example_errorref}
\end{figure}

\begin{figure}[H]
  \centering
  \includegraphics[width=\linewidth]{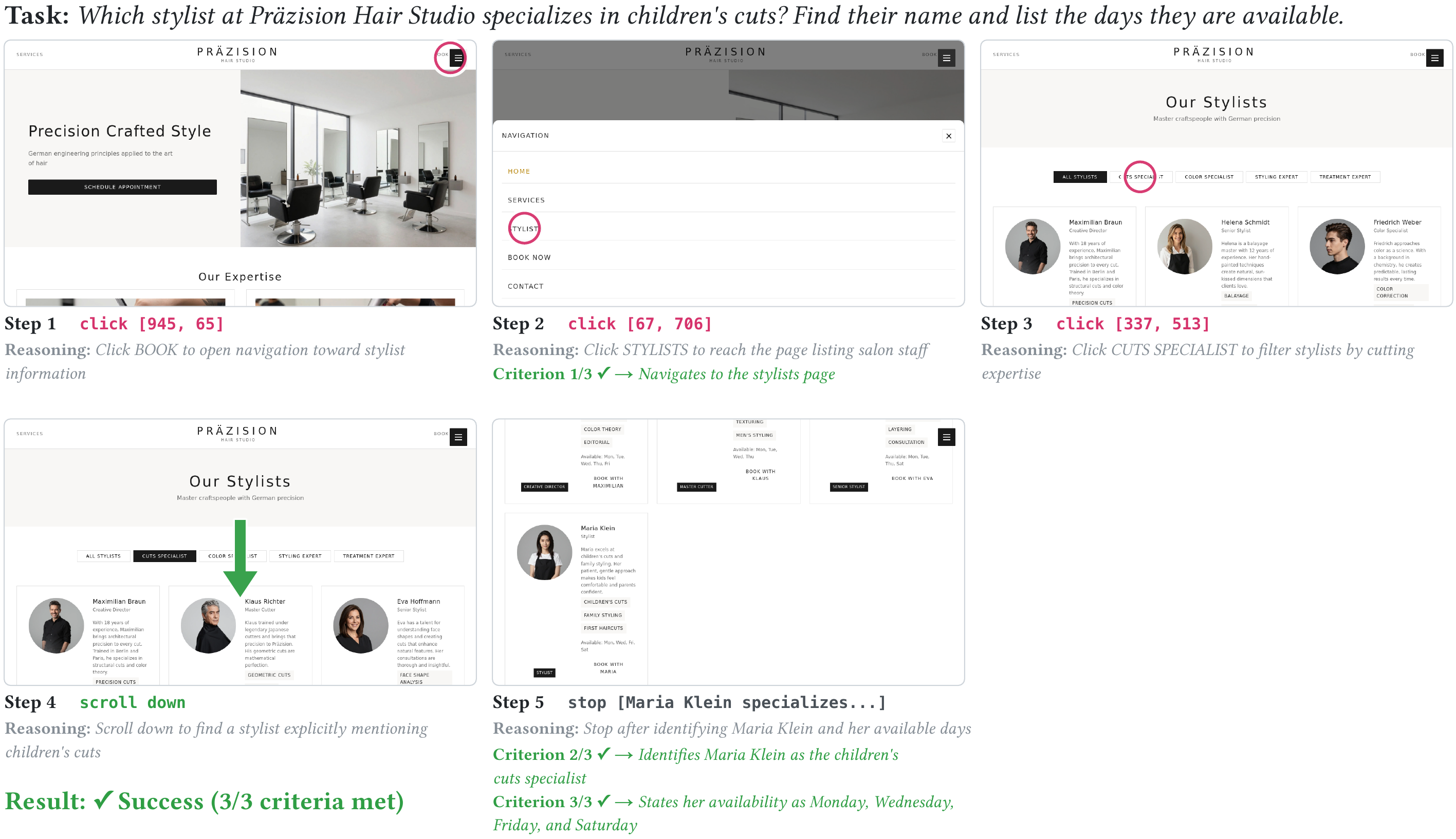}
  \caption{Example trajectory of \modelnameeightb solving a \method-val task.}
  \label{fig:app_trajectory_example_hair}
\end{figure}

\begin{figure}[H]
  \centering
  \includegraphics[width=\linewidth]{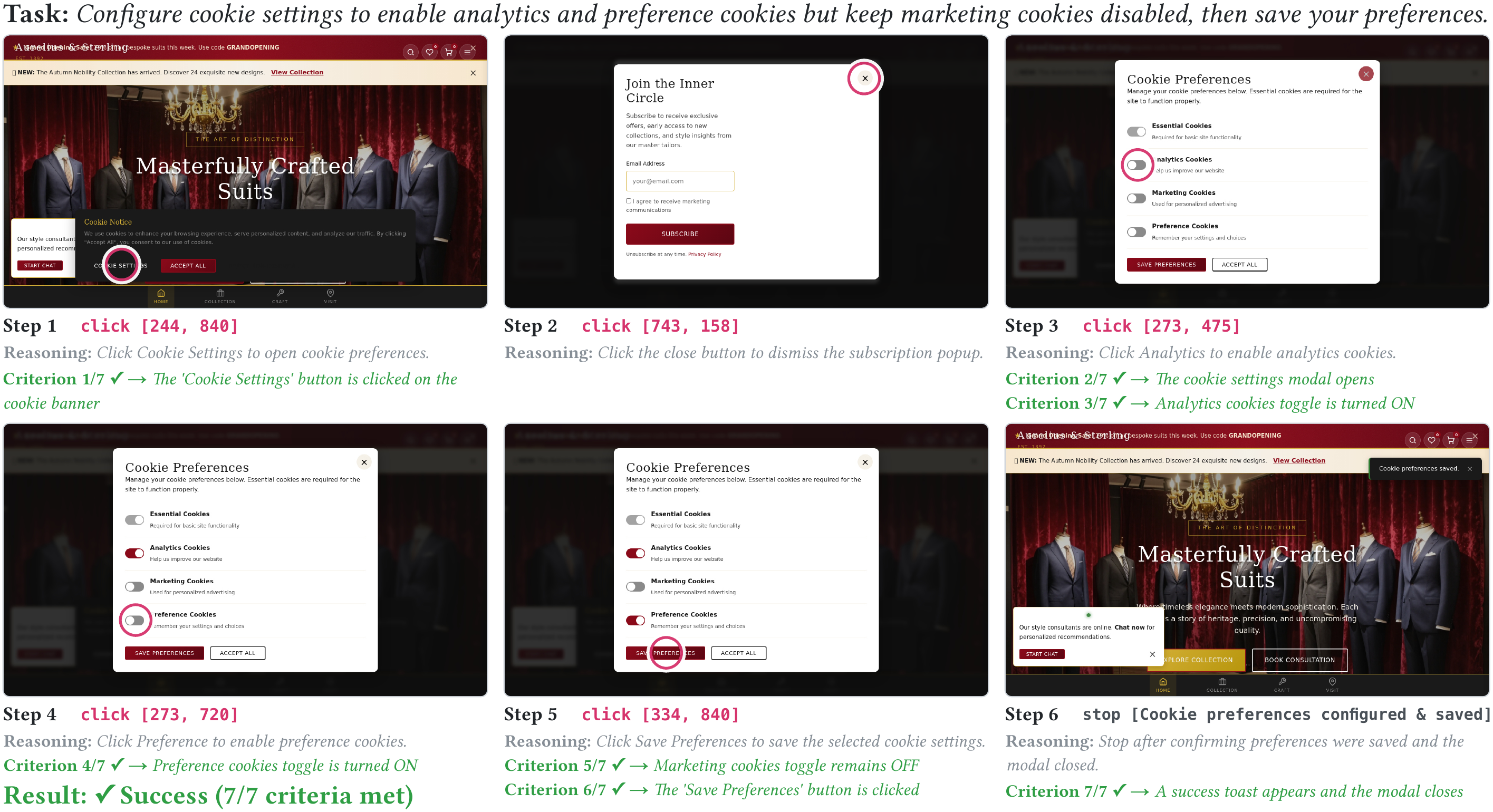}
  \caption{Example trajectory of \modelnameeightb solving a \method-val task.}
  \label{fig:app_trajectory_example_cookie}
\end{figure}

\begin{figure}[H]
  \centering
  \includegraphics[width=\linewidth]{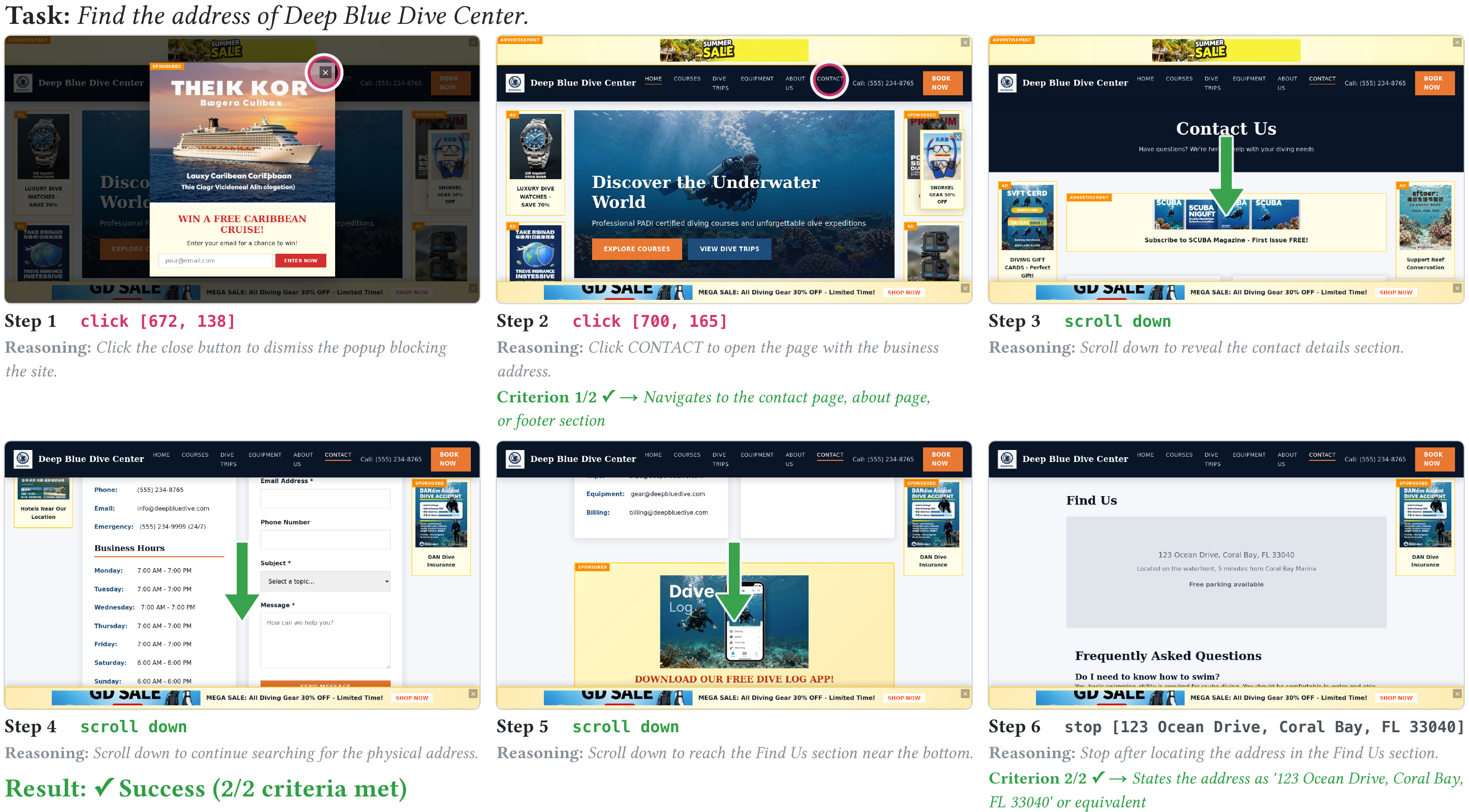}
  \caption{Example trajectory of \modelnameeightb solving a \method-val task. This synthetic website was designed to contain substantial visual clutter, such as popups and ad banners, yet the agent successfully navigates the page.}
  \label{fig:app_trajectory_example_clutter}
\end{figure}

\begin{figure}[H]
  \centering
  \includegraphics[width=\linewidth]{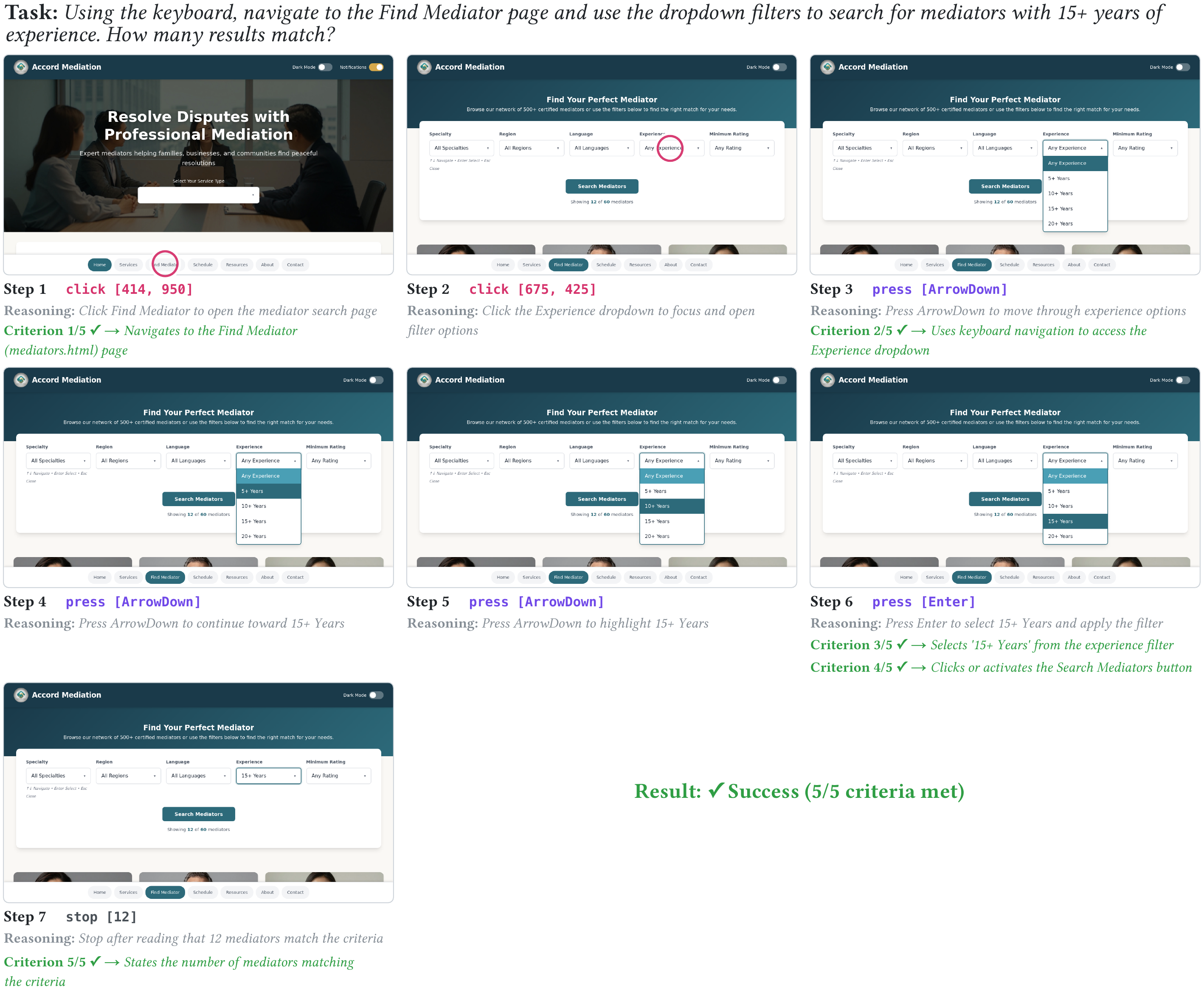}
  \caption{Example trajectory of \modelnameeightb solving a \method-val task. Although the agent could select the correct dropdown entry directly via clicking, \methodsynth tasks are designed to train web navigation capabilities broadly; in this case, the task explicitly requires completion through keyboard navigation.}
  \label{fig:app_trajectory_example_keyboardnav}
\end{figure}

\begin{figure}[H]
  \centering
  \includegraphics[width=\linewidth]{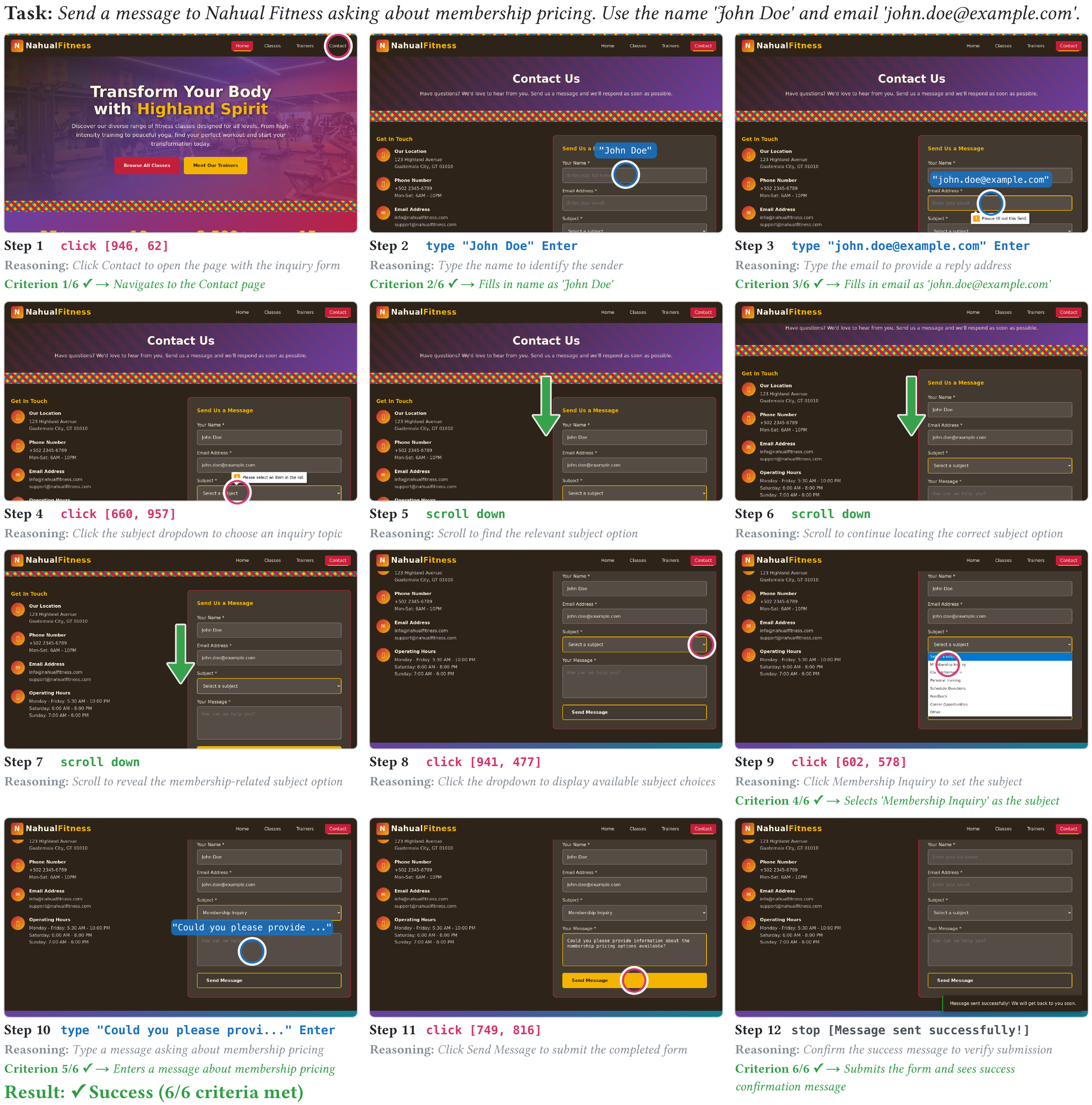}
  \caption{Example trajectory of \modelnameeightb solving a \method-val task. The agent attempts to open the dropdown menu but, as it lies near the bottom of the page, must first scroll down to reveal it.}
  \label{fig:app_trajectory_example_fitness}
\end{figure}

\begin{figure}[H]
  \centering
  \includegraphics[width=\linewidth]{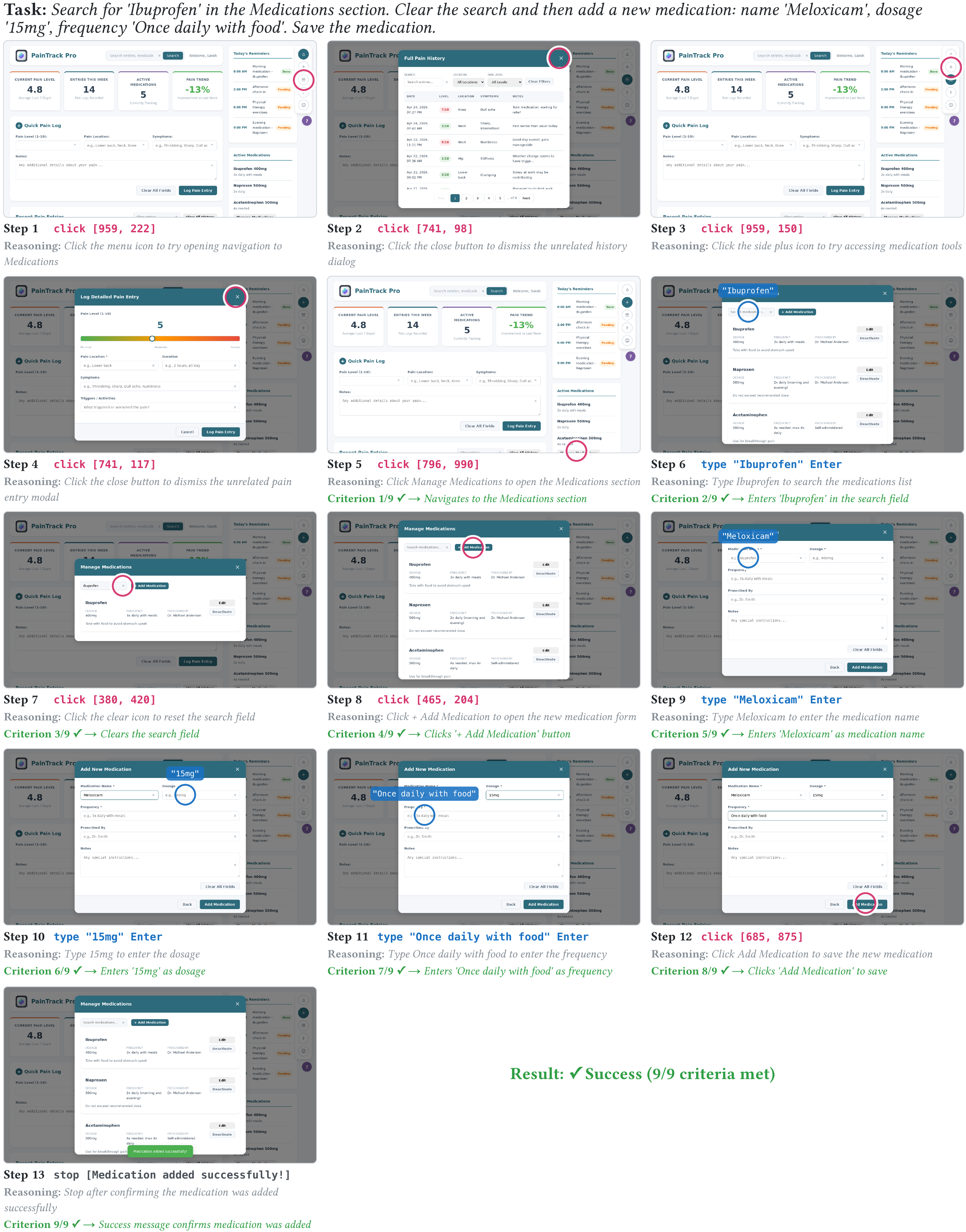}
  \caption{Example trajectory of \modelnameeightb solving a \method-val task. The agent initially opens two incorrect menus before identifying \textit{Manage Medications} at the bottom of the screen and successfully editing the medication.}
  \label{fig:app_trajectory_example_medication}
\end{figure}

In \Cref{fig:app_trajectory_example_errorref,fig:app_trajectory_example_hair,fig:app_trajectory_example_cookie,fig:app_trajectory_example_clutter,fig:app_trajectory_example_keyboardnav,fig:app_trajectory_example_fitness,fig:app_trajectory_example_medication} we show trajectory visualizations of \modelnameeightb solving \method-val tasks of various difficulties, and displaying different visual grounding and web navigation skills.

\section{Additional Training Details} \label{sec:appendix_training}                                                      
                                                                                          
\subsection{Training hyperparameters} \label{subsec:appendix_hparams}                                                                              
We performed a lightweight hyperparameter exploration for both SFT and RL stages. \Cref{tab:sft_hparams,tab:rl_hparams} list the values explored.

  \begin{table}[h]                                                                                                                                                                   
    \caption{SFT hyperparameters. \textbf{Bold} values indicate selected configurations that performed similarly.}                                                                                            
    \label{tab:sft_hparams}                                                                                                                                                          
    \centering                                                                                                                                                                       
    \begin{tabular*}{\linewidth}{@{}l@{\extracolsep{\fill}}l@{}}                                                                                                                     
      \toprule                                                                                                                                                                       
      Hyperparameter & Values \\                                                                                                                                                     
      \midrule                                                                                                                                                                       
      Base model & Qwen3-VL-8B-Instruct \\                                                                                                                                           
      Trainable components & Language model only (vision tower + projector frozen) \\                                                                                                
      Learning rate & 1e-5, \textbf{1e-6} \\                                                                                                                                         
      LR schedule & Cosine \\                                                                                                                                                        
      Warmup ratio & 0.1 \\                                                                                                                                                          
      Epochs & 1, \textbf{2}, \textbf{3} \\                                                                                                                                          
      Image resolution & 640$\times$360, \textbf{1280$\times$720} \\                                                                                                                 
      Cutoff length & 80{,}000 \\                                                                                                                                                    
      Effective batch size & 8 \\                                                                                                                                                    
      Precision & bf16 \\                                                                                                                                                            
      \bottomrule                                                                                                                                                                    
    \end{tabular*}                                                                                                                                                                   
  \end{table}                                                                                                                                                                        
                                                                                                                                                                                     
  \begin{table}[h]                                                                                                                                                                   
    \caption{RL hyperparameters. \textbf{Bold} values indicate selected configurations that performed similarly.}                                                                                             
    \label{tab:rl_hparams}                                                                                                                                                           
    \centering                                                                                                                                                                       
    \begin{tabular*}{\linewidth}{@{}l@{\extracolsep{\fill}}l@{}}                                                                                                                     
      \toprule                                                                                                                                                                       
      Hyperparameter & Values \\                                                     
      \midrule                                                                                                                                                                       
      Base model & Qwen3-VL-8B-Instruct, \textbf{SFT model} \\                                                                                                                       
      Algorithm & Dr.\ GRPO \\                                                                                                                                                       
      Learning rate & \textbf{1e-5}, \textbf{5e-6}, 1e-6 \\                                                                                                                          
      LR schedule & Constant \\                                                                                                                                                      
      KL coefficient & \textbf{0}, \textbf{0.005}, \textbf{0.01}, 0.02, 0.03, 0.05 \\                                                                                                
      Batch size & 64, \textbf{128}, \textbf{256} \\                                                                                                                                 
      Rollouts per prompt ($n$) & \textbf{4}, \textbf{8}, 16 \\                                                                                                                               
      Rounds ($R$) & 15, \textbf{25}, 35 \\                                                                                                                                          
      PPO mini-batch size & 32 \\                                                                                                                                                    
      Max grad norm & 1.0 \\                                                                                                                                                         
      Precision & bf16 \\                                                                                                                                                            
      Image resolution & 1280$\times$720 \\                                                                                                                                          
      Max model context length & 80{,}000 \\                                                                                                                                         
      Max new tokens per action & 512 \\                                                                                                                                             
      \bottomrule                                                                                                                                                                    
    \end{tabular*}                                                                                                                                                                   
  \end{table}

\subsection{Compute}                                                                                                                                                               
  \label{subsec:appendix_compute}                                                                                                                                                                         
  All SFT and RL experiments are conducted on 8 NVIDIA B200 GPUs. Environments are served locally on the same node. Each RL training step generates 1024 rollouts (256 prompts       
  $\times$ 4 rollouts per prompt), each with up to 25 interaction rounds, completing in approximately 28 minutes with our local environment setup. 

\subsection{Action space}                                                                                                                                                          
  \label{subsec:appendix_action_space}                                                                                                                                                                          
  \Cref{tab:action_space} lists the full action space used by our visual web agent.                                                                                             
\begin{table}[h]                                                                                                                                                                   
    \centering                                                                       
    \caption{Action space of visual web agent. Coordinate-based actions take pixel positions as arguments.}                                                                          
    \label{tab:action_space}                                                                                                                                                         
    \small                                                                                                                                                                           
    \begin{tabular*}{\linewidth}{@{}l@{\extracolsep{\fill}}l@{}}                                                                                                                     
      \toprule                                                                                                                                                                       
      \textbf{Action} & \textbf{Description} \\                                                                                                                                      
      \midrule                                                                                                                                                                       
      \texttt{click(x, y)} & Click at pixel coordinates \\                                                                                                                           
      \texttt{hover(x, y)} & Hover at pixel coordinates \\                                                                                                                           
      \texttt{type(text, [x, y], [enter])} & Type text, optionally at coordinates and press enter \\                                                                                 
      \texttt{press(key)} & Press a keyboard key \\                                                                                                                                  
      \texttt{scroll(direction, [amount])} & Scroll in a given direction \\                                                                                                           
      \texttt{go\_back()} & Navigate back in history \\                                                                                                                              
      \texttt{go\_forward()} & Navigate forward in history \\                                                                                                                        
      \texttt{wait()} & Wait for page to load \\                                                                                                                                     
      \texttt{stop(response)} & Submit response and end episode \\                                                                                                                   
      \bottomrule                                                                                                                                                                    
    \end{tabular*}                                                                                                                                                                   
  \end{table}  

\subsection{Training reward judge prompt}\label{subsec:appendix_judge_prompt}                                               
       
  During RL training, we use a VLM-based judge (GPT-4o) to evaluate whether the agent successfully completed each task. The judge receives the full trajectory (screenshots and      
  actions) and evaluates against task-specific criteria. Samples labeled as ``website failure'' are discarded from the training batch. The prompt is shown below.

\begin{tcolorbox}[colback=gray!5, colframe=gray!50, title=Training Judge Prompt, fonttitle=\small\bfseries, boxrule=0.4pt, left=4pt, right=4pt, top=4pt, bottom=4pt]               
  \small\ttfamily                                                                    
  \textbf{System:}\\                                                                                                                                                                 
  You are an expert evaluator for web navigation tasks.\\[4pt]                                                                                                                       
  Analyze the provided trajectory steps and determine if the agent successfully completed the following task: \{task\}\\[4pt]                                                        
  Respond with one of the following based on the task-specific criteria below:\\                                                                                                     
  - `correct' -- The agent successfully completed the task\\                                                                                                                         
  - `incorrect' -- The agent failed to complete the task due to its own mistakes\\                                                                                                   
  - `website failure' -- The agent was making reasonable progress but was blocked by technical issues beyond its control\\[4pt]                                                      
  Task-specific evaluation criteria: \{criteria\}\\[4pt]                                                                                                                             
  Technical issues that qualify for `website failure':\\                                                                                                                             
  - Page timeouts or loading failures\\                                                                                                                                              
  - Blank or empty pages that fail to render\\                                                                                                                                       
  - Connection errors or server errors (5xx responses)\\                                                                                                                             
  - CAPTCHA or bot detection blocking the agent\\                                                                                                                                    
  - Pages stuck in infinite loading states\\                                                                                                                                         
  - Elements that fail to become interactive despite multiple attempts\\[4pt]                                                                                                        
  To determine if a website issue occurred, look for these indicators in the trajectory:\\                                                                                           
  - The agent repeatedly tries the same reasonable action without success\\                                                                                                          
  - Screenshots show loading spinners, error messages, or blank content\\                                                                                                            
  - The agent uses `wait' actions multiple times without page progress\\                                                                                                             
  - The agent's actions are correct but the page state doesn't change as expected\\[4pt]                                                                                             
  Important: Only use `website failure' when the agent was on a reasonable path toward completing the task. If the agent made fundamental mistakes before encountering technical     
  issues, respond with `incorrect'.\\[4pt]         

  These are the actions the agent can take:\\
  \textit{[Action space description omitted for brevity; see \Cref{tab:action_space}]}\\[8pt]                                                                                   
  \textbf{User:}\\                                                                                                                                                                   
  Step 1 - Screenshot: \textit{[image]}\\                                                                                                                                            
  Step 1 - Agent Action: \textit{[action]}\\                                                                                                                                         
  Step 2 - Screenshot: \textit{[image]}\\                                                                                                                                            
  Step 2 - Agent Action: \textit{[action]}\\                                                                                                                                         
  \dots\\[4pt]                                                                                                                                                                       
  Based on these trajectory steps, did the agent successfully complete the task? First respond with your decision followed by your reasoning.                                        
  \end{tcolorbox} 

\begin{figure}[h]                                                                                                                                                                  
    \centering                                                                       
    \includegraphics[width=0.6\linewidth]{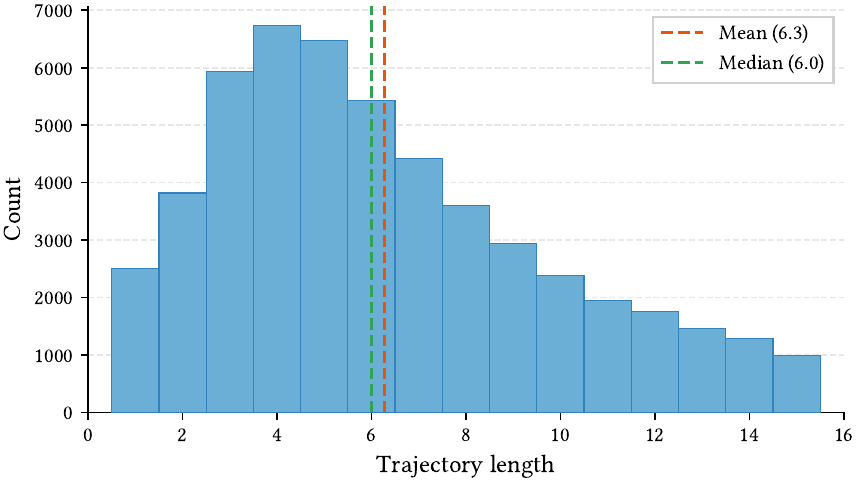}                                                                                                    
    \caption{Distribution of trajectory lengths in the SFT training data.}                                               
    \label{fig:sft_traj_hist}                                                                                                                                                        
  \end{figure}  

\section{Data Statistics}

\subsection{Environments}\label{sec:appendix_data_stats}
\begin{figure}[t]
  \centering
  \includegraphics[width=\linewidth]{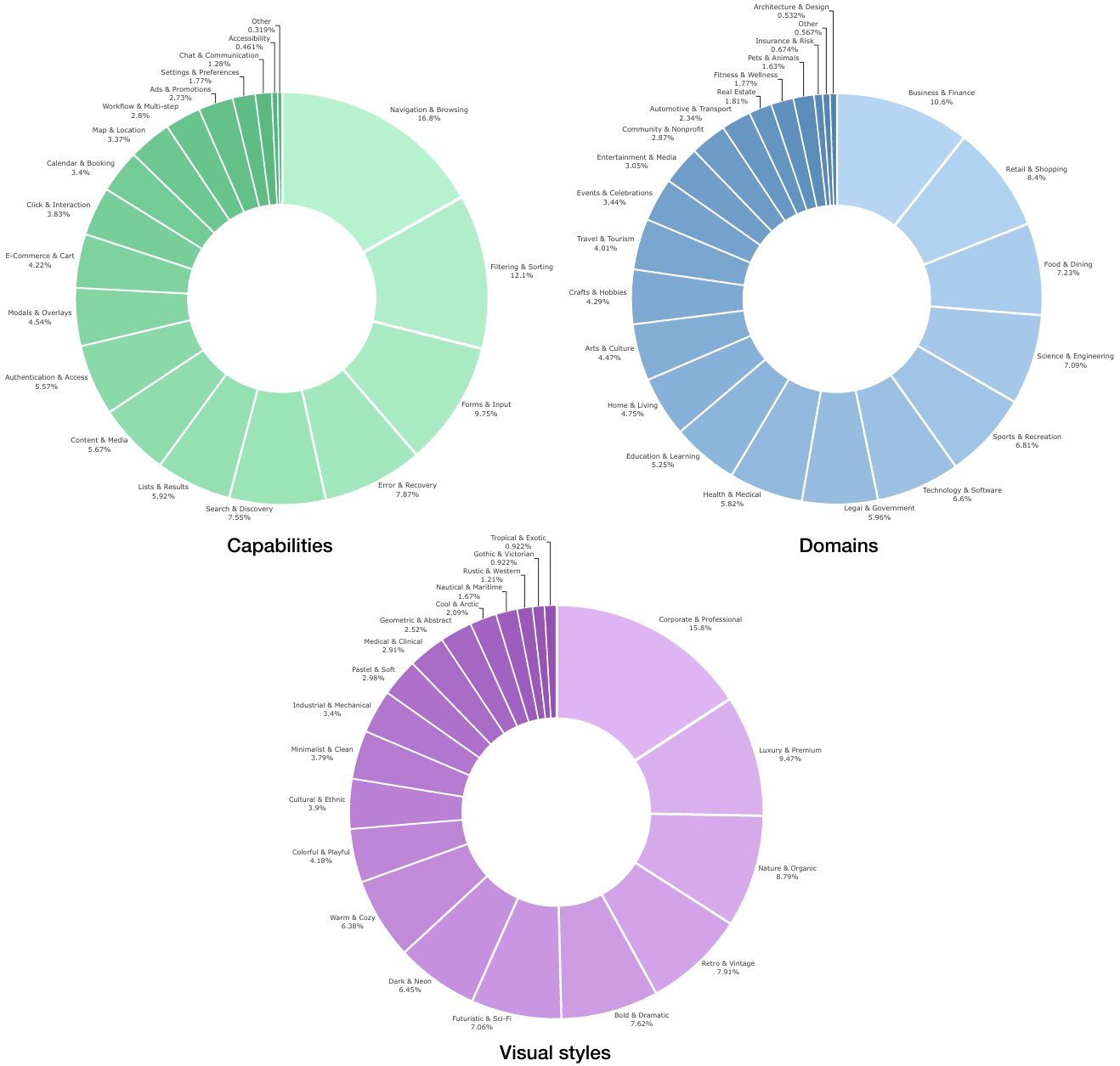}
  \caption{\textbf{\methodsynth site grouping}: Distribution of synthetic web browsing tasks grouped across capability categories (top left), domain categories (top right), and visual style categories (bottom).}
  \label{fig:app_om2w_distribution}
\end{figure}

\Cref{fig:app_om2w_distribution} shows the distribution of synthesized environments across capability, domain, and visual style categories.

\subsection{SFT Trajectories}

  \Cref{fig:sft_traj_hist} shows the distribution of trajectory lengths across the 51.7K SFT training trajectories.

\end{document}